%% file: main.tex
\documentclass[10pt,twocolumn,letterpaper]{article}

\usepackage{iccv}
\usepackage{times}
\usepackage{epsfig}
\usepackage{graphicx}
\usepackage{amsmath}
\usepackage{amssymb}
\usepackage{multirow}
\usepackage{booktabs,tabularx}
\usepackage[margin=1in]{geometry} % set margins to meet your document's needs
\newcolumntype{Y}{>{\raggedright\arraybackslash}X} 

\newcommand{\epx} {\mathbb{E}_{\mathbf{x}\sim P_{\mathbf{x}},\mathbf{y}\sim P_{\mathbf{y}}}}
\newcommand{\tx}{\tilde{\mathbf{x}}}
\newcommand{\x}{\mathbf{x}}
\newcommand{\y}{\mathbf{y}}
\newcommand{\M}{\mathbf{M}}

\usepackage[breaklinks=true,bookmarks=false]{hyperref}

\iccvfinalcopy % *** Uncomment this line for the final submission

\def\iccvPaperID{6414} % *** Enter the ICCV Paper ID here

\ificcvfinal\fi

\begin{document}

%%%%%%%%% TITLE
\title{Born Identity Network:\\ Multi-way Counterfactual Map Generation to Explain a Classifier's Decision}

\author{
Kwanseok Oh$^{1,}$\thanks{These authors equally contributed to this work.} \quad Jee Seok Yoon$^{2,}$\textsuperscript{*} \quad Heung-Il Suk$^{1,2,}$\thanks{Corresponding author.} \\
$^1$Department of Artificial Intelligence, Korea University, Korea\\
$^2$Department of Brain and Cognitive Engineering, Korea University, Korea\\
{\tt\small  \{ksohh, wltjr1007, hisuk\} @korea.ac.kr} \\
}

\maketitle
% Remove page # from the first page of camera-ready.
\ificcvfinal\thispagestyle{empty}\fi

%%%%%%%%% ABSTRACT
\begin{abstract}
    There exists an apparent negative correlation between performance and interpretability of deep learning models. In an effort to reduce this negative correlation, we propose a Born Identity Network (BIN), which is a post-hoc approach for producing multi-way counterfactual maps. A counterfactual map transforms an input sample to be conditioned and classified as a target label, which is similar to how humans process knowledge through counterfactual thinking. For example, a counterfactual map can localize hypothetical abnormalities from a normal brain image that may cause it to be diagnosed with a disease. Specifically, our proposed BIN consists of two core components: Counterfactual Map Generator and Target Attribution Network. The Counterfactual Map Generator is a variation of conditional GAN which can synthesize a counterfactual map conditioned on an arbitrary target label. The Target Attribution Network provides adequate assistance for generating synthesized maps by conditioning a target label into the Counterfactual Map Generator. We have validated our proposed BIN in qualitative and quantitative analysis on MNIST, 3D Shapes, and ADNI datasets, and showed the comprehensibility and fidelity of our method from various ablation studies. Code is available at: \url{https://github.com/ksoh97/BIN}.
\end{abstract}

%%%%%%%%% BODY TEXT
\section{Introduction}
    As deep learning has shown its success in various domains, there has been a growing need for interpretability and explainability in deep learning models. The black-box nature of deep learning models limits their real-world applications in fields, especially, where fairness, accountability, and transparency are essential. Moreover, from the end-user point of view, it is a crucial process that requires a clear understanding and explanation at the level of human knowledge. However, achieving high performance with interpretability is still an unsolved problem in the field of explainable AI (XAI) due to their apparent negative correlation~\cite{hooker2019benchmark} (\ie interpretable models tend to have a lower performance than black-box models).
    
    \begin{figure}[t]
    \centering
    {\includegraphics[width=1.\columnwidth]{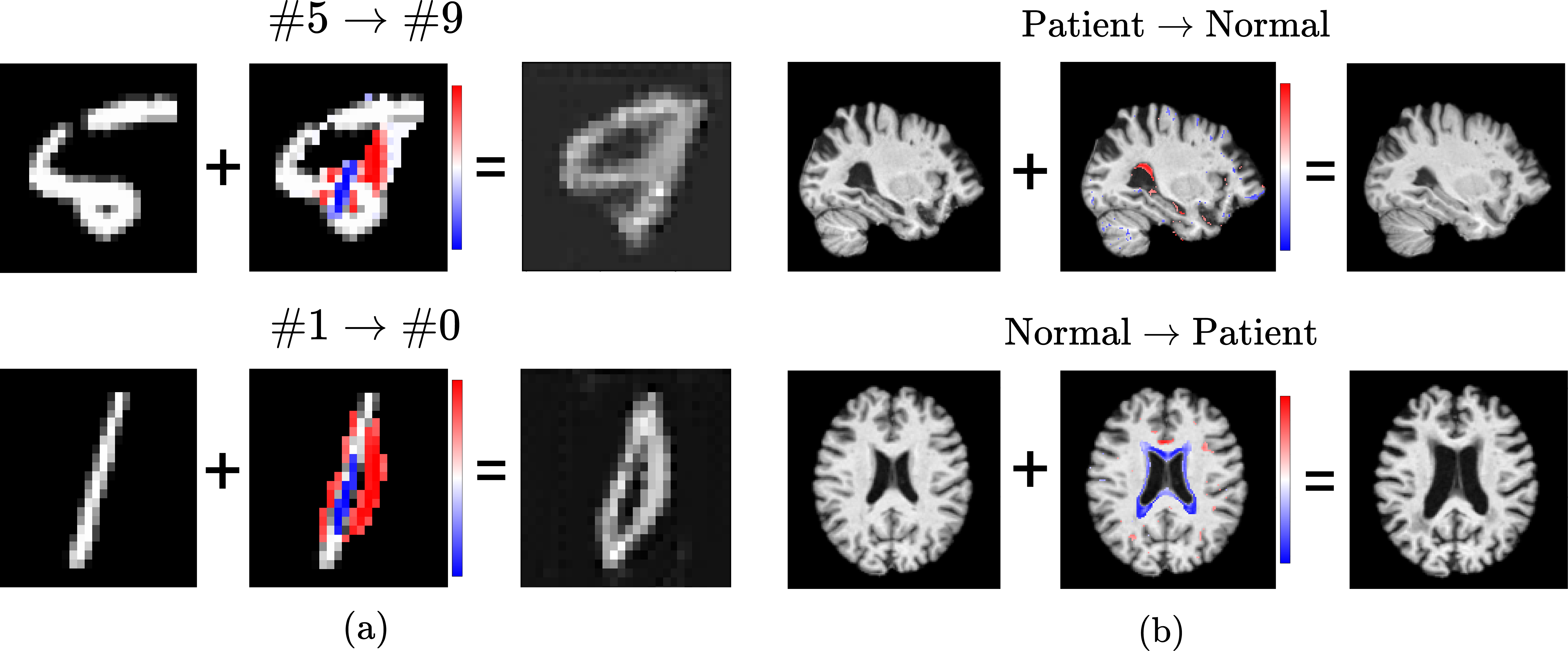}
    \caption{We propose an approach for producing counterfactual maps as a step towards counterfactual reasoning, which is a process of producing hypothetical realities given observations. For example, a counterfactual map can localize hypothetical, yet very possible, abnormalities from a brain image of a normal subject that may cause the subject to be diagnosed with a disease.}\label{fig:abstract}}
    \end{figure}
    
    Reducing this negative correlation in performance and interpretability has been a long-standing goal in the field of XAI. In the early era of XAI~\cite{gilpin2018explaining}, researchers have proposed various methods for discovering or identifying the regions that have the most influence on deriving the outcome of the classifier~\cite{selvaraju2017grad,smilkov2017smoothgrad,sundararajan2017axiomatic,bach2015pixel,montavon2017explaining,shrikumar2017learning,simonyan2013deep,zeiler2014visualizing}. The main objective of these early era XAI methods is to answer {\em why} and {\em how} a model has made its decision. However, recent XAI methods induce to answer the question that can offer a more fundamental explanation: ``What are the \textit{hypothetical alternative scenarios} that may have altered the outcome?''. This sort of explanation is defined at the root of {\em counterfactual reasoning}. Counterfactual reasoning can provide an explanation at the level of human knowledge since it can explain a model’s decision in hypothetical situations. Thus, we propose to show a higher-level visual explanation of a deep learning model similar to that of how humans process knowledge, \ie, through the means of a counterfactual map.
    
    A {\em counterfactual map} is a map that can transform an input sample that was originally classified as one label to be classified as another. For example, a counterfactual map transforms a digit image to be an image of another number (Figure \ref{fig:abstract}-(a)). This counterfactual map explains what kind of structural changes are required for a digit image to be another number with the least amount of modification to the original image. A real-world application can be of a medical image analysis of the brain (Figure \ref{fig:abstract}-(b)), where a counterfactual map would describe which Region Of Interests (ROIs) may cause a normal subject to be diagnosed with a disease (also with least amount of modification to the original image).

    %1. medical motivation, 2. multi-way motivation
    % progression -> 환자의 질병 진행 상태와 척도 / 계층화 (stratification)==현재 이 환자가 어떤 증상에 가까운지 /
    
    % It is also noteworthy that from a clinical point of view,the process of identifying a patient’s status is mostly  a multi-class classification task, differentiating among a set of diseases that cause similar symptoms, rather than just separating between normal and abnormal. In this sense, our multi-way counterfactual map generation method can be an impactful tool for more generic clinical circumstances.

    % {\modification Distinguishing these diseases provides great benefits to the clinician, and increases the chances of enabling specific treatments to improve the patient's condition and save significant medical costs~\cite{hunter2015medical}.} Moreover, the counterfactual map can support a more precise diagnosis by identifying significant neuropathological changes~\cite{lindberg2012hippocampal,gerardin2009multidimensional}, as well as potential biomarker regions that are not easily noticed by clinicians. Thus, the motivation of this paper is not only to interpret the decisions of classifiers that derive high performance in various domains but also to provide diagnostic auxiliary information to clinicians, as its real-world application.

    To our knowledge, most works on producing counterfactual explanations are generative models. Most notably, these works utilize Generative Adversarial Network (GAN) and its variants~\cite{goyal2019explaining,chang2018explaining,van2019interpretable,dash2020counterfactual,sauer2021counterfactual} to synthesize a counterfactual explanation. Although these works can generate meaningful counterfactual explanations, two fundamental problems limit their application in the real-world. First, generative models are a typical example of models that experience the aforementioned negative correlation between performance and interpretability. To address this issue, we propose a method that can produce counterfactual maps from a pre-trained model
    %. This post-hoc nature of our work not only provides a generalized framework that can be applied to most neural networks but also reduces the negative correlation since it can be applied to a model that already has a high performance 
    (\ie, we can focus solely on interpretability with the higher performance given beforehand).
    
    Second, recent works on counterfactual explanation can only produce a single-\cite{dash2020counterfactual,goyal2019explaining,chang2018explaining} or dual-\cite{goyal2019counterfactual,Wang_2020_CVPR} way explanation. In other words, they only consider one or two hypothetical scenarios for counterfactual reasoning (\eg, single-way counterfactual map can only transform a normal subjects to Alzheimer's disease patients, and vice versa for dual-way maps). With the help of a target attribution mechanism, our work, to the best of our knowledge, is the first to propose a \textit{multi-way} counterfactual reasoning (\eg, producing maps that can freely transform any image samples to or from normal subject, mild cognitive impairment patient, and Alzheimer's disease patient).
    
    The motivation behind a \textit{multi-way} counterfactual map is (in contrast to single-/dual-way) the stratification of hypothetical alternatives, especially in the medical domain. A single-/dual-way counterfactual map is only able to justify the outcome in one or two extreme tails (\eg, disease or normal), while in real-world, diseases are usually stratified by the severity and/or progression. A multi-way counterfactual map provides a natural proxy for stratification of diseases by providing hypothetical scenarios for prodromal stages (\ie, preceding stages) of a disease.
    
    To this end, we propose Born Identity Network (BIN) that produces a counterfactual map using two components: The Counterfactual Map Generator (CMG), and the Target Attribution Network (TAN). The CMG is a variant of conditional GAN~\cite{mirza2014conditional} that synthesizes a conditioned map, while the TAN provides adequate assistance to the CMG by enforcing a target counterfactual attribute to the synthesized conditioned map. We have validated our proposed BIN in qualitative and quantitative analysis on MNIST, 3D Shapes, and ADNI datasets, and showed the comprehensibility and fidelity of our method from various ablation studies.
    
    % To evaluate our proposed framework, we perform a suite of analyses from the perspective in various data domains: MNIST, 3D Shapes, and {\js ADNI dataset}. First, we demonstrate and analyze the counterfactual maps for these above datasets, as the qualitative evaluation for our work. Second, to quantitatively validate the counterfactual map, we further calculate a correlation score between the counterfactual map and its ground-truth map. Finally, we examine how each component of BIN works toward creating a counterfactual map with a suite of exhaustive ablation studies. 
    
    The main contributions of our study are as follows:
    \begin{itemize}
        \item We propose Born Identity Network (BIN\footnote { We have coined this word to emphasize that the produced counterfactual map tries to maintain the identity of the original input sample while changing only the slightest ROIs absolutely necessary for the target class.}), which, to the best of our knowledge, is the first work on producing counterfactual reasoning in \textit{multiple} hypothetical scenarios.
        \item {A \textit{Multi-way} counterfactual map can provide a more precise explanation which accounts for severity and/or progression of a disease.}
        \item BIN is a generalized interpretation model which can be plugged-in to most pretrained neural networks without any modification to the weights of the pretrained model.
    \end{itemize}
    
\section{Related Works}
    In this section, we describe various works proposed for explainable AI (XAI). First, we divide XAI into a general framework of attribution-based explanation and a more recent framework of counterfactual explanation. Second, we have compared different approaches to counterfactual explanation in the Supplementary.
    
\subsection{Attribution-based Explanations}
     Attribution-based explanation refers to discovering or identifying the regions that have the most influence on deriving the outcome of a model. These approaches can further be separated into the gradient-based and reference-based explanation. First, the gradient-based explanation highlights the activation nodes that most contributed to the model’s decision. For example, Class Avtivation Map (CAM)~\cite{zhou2016learning}, and Grad-CAM~\cite{selvaraju2017grad} highlight activation patterns of weights in a specified layer. In a similar manner, DeepTaylor~\cite{montavon2017explaining}, DeepLift~\cite{shrikumar2017learning}, and Layer-wise Relevance Propagation (LRP)~\cite{bach2015pixel} highlight the gradients with regard to the prediction score. These approaches usually suffer from vanishing gradients due to the ReLU activation, and Integrated Gradients~\cite{sundararajan2017axiomatic} resolves this issue through sensitivity analysis. However, a crucial drawback of attribution-based approaches is that they tend to ignore features with relatively low discriminative power or highlight only those with overwhelming feature importance. Second, the reference-based explanation~\cite{zeiler2014visualizing,fong2017interpretable,chang2018explaining,dhurandhar2018explanations} focuses on changes in model output with regards to perturbation in input samples. Various perturbation methods, such as masking~\cite{dabkowski2017real}, heuristic~\cite{chang2018explaining} (\eg, blurring, and random noise), region of the distractor image as reference for perturbation~\cite{goyal2019counterfactual}, or synthesized perturbation~\cite{dhurandhar2018explanations,zeiler2014visualizing,fong2017interpretable}, has been proposed. One general drawback of these aforementioned attribution-based explanations is that they tend to produce low-resolution and blurred salient map. In contrast, our work can produce a crisp salient map with the same resolution as the input sample using a GAN.
    
\subsection{Counterfactual Explanations}
     Recently, more researchers have focused on counterfactual reasoning as a form of higher-level explanation. Counterfactual explanation refers to analyzing the model's output with regards to hypothetical scenarios. For example, a counterfactual explanation could highlight those regions that may (hypothetically) cause a normal subject to be diagnosed with a disease. VAGAN~\cite{baumgartner2018visual} uses a variant of GAN that synthesizes a counterfactual map that transforms an input sample to be classified as another label. However, VAGAN can only perform single-way synthesis (\eg, the map that transforms input originally classified as A to be classified as B, but not vice versa), which limits the counterfactual explanation to also become single-sided. ICAM~\cite{bass2020icam} was proposed as an extension to VAGAN to produce a dual-way counterfactual explanation. However, most real-world applications cannot be explained through single-/dual-way explanations due to their complexity. Thus, our work proposes an approach for producing multi-way counterfactual explanations. %(\eg, counterfactual map that can transform a digit image to be classified as any other number).
     By presenting the validity of the method we propose for a multi-way explanation on the medical domain, it is expected that our proposed method can be one of the milestones towards the explanation of multi-class classification in neurodegenerative disease diagnosis.

\begin{figure}[t]
    \centering
    {\includegraphics[width=0.7\columnwidth]{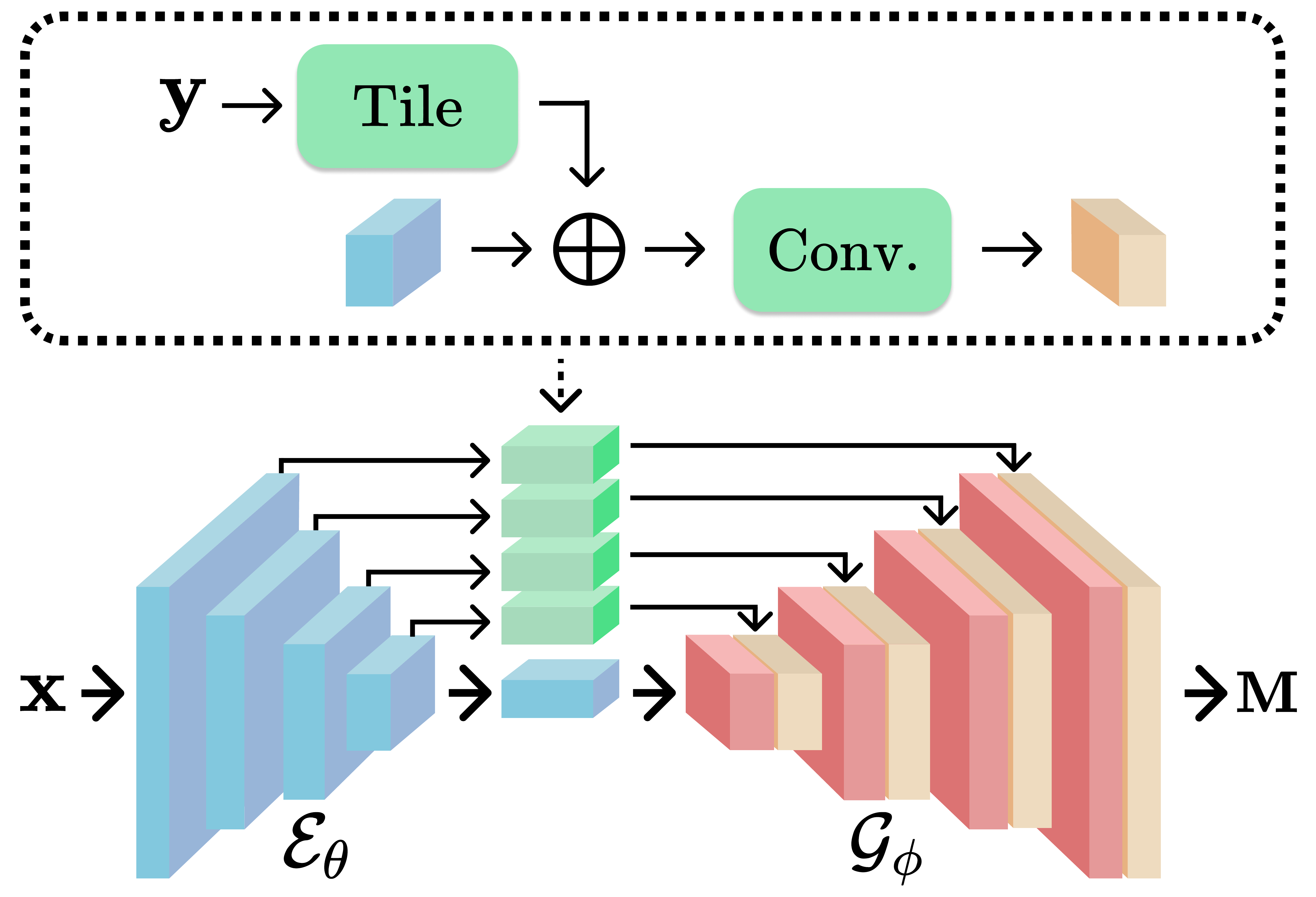}}
    \caption{A detailed view of the Counterfactual Map Generator (CMG). A tiled one-hot target label $\y$ is concatenated to the skip connection. This enables the CMG to condition the counterfactual maps to be conditioned on an arbitrary target condition.}\label{fig:Detail of CFmap generator}
\end{figure}

\begin{figure*}[t]
\centering
\includegraphics[width=1.\linewidth]{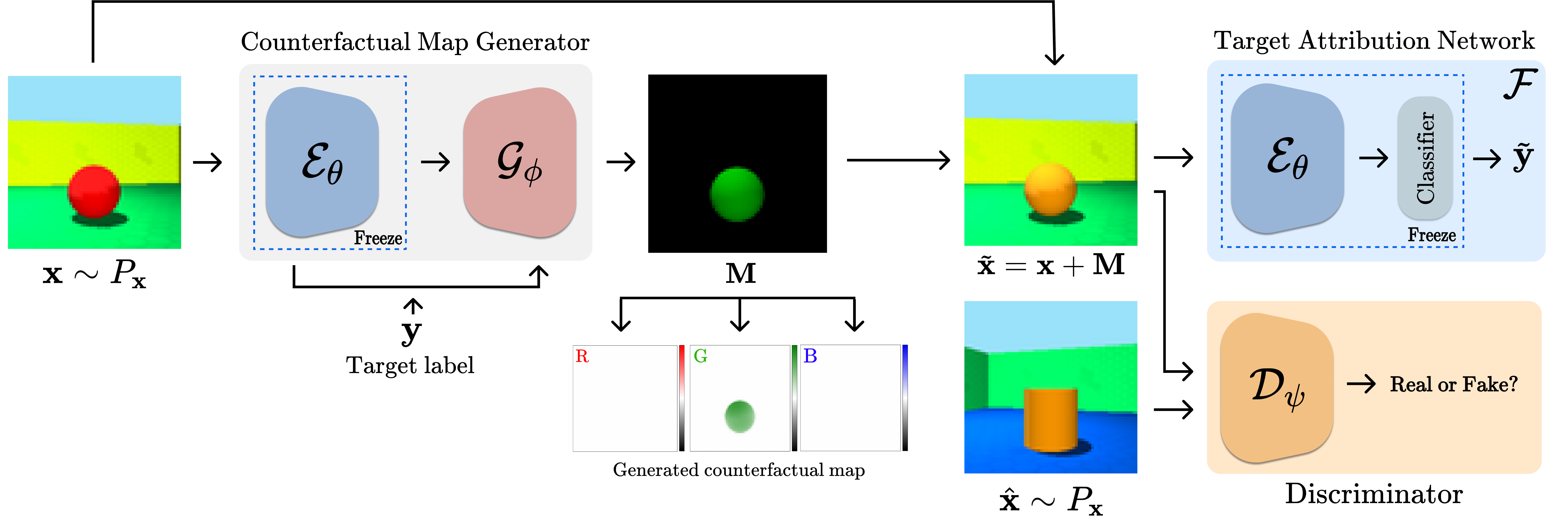}
\caption{Schematic overview of Born Identity Network (BIN). There are two major components of BIN: Counterfactual Map Generator (CMG) and Target Attribution Network (TAN). The CMG synthesized a counterfactual map conditioned on arbitrary target label, while the TAN work towards enforcing target label attributes to the synthesized map.}
\label{fig: Overall framework}
\end{figure*}

\section{Multi-way Counterfactual Map}
    Here, we formally define a multi-way counterfactual map. First, we define the dataset $P=(P_\mathbf{x},P_\mathbf{y})$, where $(\mathbf{x},\mathbf{y})\sim P_{\mathbf{x},\mathbf{y}}$ denote the data, label pair and $\mathbf{x}\sim P_\mathbf{x}, \mathbf{y}\sim P_\mathbf{y}$ denote the unpaired data and label. We define a counterfactual map as a salient map that is able to produce a counterfactual explanation of a classifier $\mathcal{F}$. Formally, a single- or dual-way counterfactual map $\mathbf{M}_\mathbf{x}$ is a map that when added to an input sample $\mathbf{x}$, \ie, $\tilde{\mathbf{x}}=\mathbf{x}+\mathbf{M}_\mathbf{x}$, the classifier classifies it as another label complement to its original label, \eg, $\mathcal{F}(\mathbf{x})=true$ and $\mathcal{F}(\tilde{\mathbf{x}})=false$, and vice versa for dual-way counterfactual map. Here, if we are able to condition this counterfactual map with an attribute (or label), \ie, the multi-way counterfactual map $\mathbf{M}_{\mathbf{x},\mathbf{y}}$, we can transform an input sample as any other attribute of interest:
    \begin{equation}
        \tilde{\mathbf{x}}:=\mathbf{x}+\mathbf{M}_{\mathbf{x},\mathbf{y}}\quad\text{such that}\; \mathcal{F}(\tilde{\mathbf{x}})=\mathbf{y},
    \end{equation}
    where data $\mathbf{x}$ and label $\mathbf{y}$ may be unpaired, \ie, we may condition the target label to be any one-hot vector (\eg, $\mathbf{y}$ can be a random one-hot vector or the posterior probability of the original input ($\mathbf{y}'$ in Eq.~\ref{eq_cyc})).

\section{Born Identity Network}
    The goal of BIN is to induce counterfactual reasoning dependent on the target condition from a pre-trained model. To achieve this goal, we've devised BIN with two core modules, \ie, Counterfactual Map Generator (CMG) and Target Attribution Network (TAN), which work in a complementary manner in producing a multi-way counterfactual map. Specifically, the CMG synthesize conditioned maps, while the TAN enforces a target attribution to synthesized maps (Figure~\ref{fig: Overall framework}).
    
\subsection{Counterfactual Map Generator}
    The CMG is a variant of Conditional GAN~\cite{mirza2014conditional} that can synthesize a counterfactual map conditioned on a target label $\y$, \ie, $\M_{\x,\y}$. Specifically, it consists of an encoder $\mathcal{E}_\theta$, a generator $\mathcal{G}_\phi$, and a discriminator $\mathcal{D}_\psi$. First, the network design of the encoder $\mathcal{E}_\theta$ and the generator $\mathcal{G}_\phi$ is a variation of U-Net~\cite{ronneberger2015u} with a tiled target label concatenated to the skip connections (Figure~\ref{fig:Detail of CFmap generator}). This generator design enables the generation to synthesize target conditioned maps such that multi-way counterfactual reasoning is possible. As a result, the counterfactual map is formulated as following:
    \begin{equation}
        \M_{\x,\y}=\mathcal{G}_\phi\left(\mathcal{E}_\theta(\x), \y\right),
    \end{equation}
    \begin{equation}
        \tx=\x+\M_{\x,\y}.
    \end{equation}
    Finally, following a general adversarial learning scheme, the discriminator discriminates real samples $\hat{\x}\sim P_\x$ from synthesized samples $\tilde{\x}$.
    
    Counterfactual reasoning requires a good balance between the proposed hypothetical and given reality. In the following sections, we define the loss functions that guide the CMG to produce a well-balanced counterfactual map.

\subsubsection{Adversarial Loss}
    For the adversarial loss functions, we have adopted the Least Square GAN (LSGAN)~\cite{mao2017least} objective function due to its stability during adversarial training. More specifically, LSGAN objective function contributes to a stable model training by penalizing samples far from the discriminator's decision boundary. This objective function is an important choice for the BIN since the generated counterfactual maps should neither destroy the input appearance nor ignore the target attribution, \ie, it should contain a good balance between real and fake samples. To this end, the discriminator and the generator loss is defined as follows, respectively:
    \begin{align}
        \begin{split}
        \mathcal{L}_{adv}^{\mathcal{D}_{\psi}} &= \frac{1}{2}\mathbb{E}_{\hat{\mathbf{x}}\sim P_{\mathbf{x}}}[(D_{\psi}(\hat{\mathbf{x}})-1)^2] \\
         &+ \frac{1}{2}\epx[(D_{\psi}(\tx))^2],
        \end{split}
        \\
        \mathcal{L}_{adv}^{\mathcal{G}_{\phi}} &= \frac{1}{2}\epx[(D_{\psi}(\tx) - 1)^2],
    \end{align}
    where $\mathbf{\tilde{x}}=\mathbf{x}+\mathbf{M}_{\mathbf{x}, \mathbf{y}}$, $\mathbf{y}$ denote the random target label.

\subsubsection{Cycle Consistency Loss}
    The cycle consistency loss is used for producing better \textit{multi-way} counterfactual maps.
    In the scenario of an single-way counterfactual map generator, real samples $\hat{\x}$ and synthesized samples $\tilde{\x}$ are always sampled from one same specific class. This setting does not require a cycle consistency loss since the real and fake samples always have specified labels. However, since our discriminator only classifies the real or fake samples, it does not have the ability to guide the generator to produce multi-way counterfactual maps. Thus, we add a cycle consistency loss where the forward cycle produces a map with an arbitrary condition, \ie, $\M_{\x,\y}$ generated from an unpaired data $\x\sim P_\mathbf{X}$ and random target label $\mathbf{y}\sim P_{\mathbf{y}}$, and the backward cycle produces a map conditioned on the posterior probability obtained from the classifier $\mathcal{F}$:
    \begin{gather}
        \begin{aligned}
            \mathcal{L}_{cyc} = \mathbb{E}_{\mathbf{x}\sim P_{\mathbf{x}}, \mathbf{y}\sim P_{\mathbf{y}}}[\|(\Tilde{\mathbf{x}}+ \mathbf{M}_{\tilde{\mathbf{x}}, \mathbf{y}'}) - \mathbf{x}\|_1],
        \end{aligned}
        \label{eq_cyc}
    \end{gather}
    % \begin{gather}
    %     \begin{aligned}
    %         \mathcal{L}_{cyc} = \mathbb{E}_{(\mathbf{x},\mathbf{y}')\sim P_{\mathbf{x,y}},   \,\mathbf{y}\sim P_{\mathbf{y}}}[\|(\Tilde{\mathbf{x}}+ \mathbf{M}_{\tilde{\mathbf{x}}, \mathbf{y}'}) - \mathbf{x}\|_1],
    %     \end{aligned}
    % \end{gather}
    where $\mathbf{\tilde{x}}=\mathbf{x}+\mathbf{M}_{\mathbf{x}, \mathbf{y}}$, $\mathbf{y}'=\mathcal{F}(\mathbf{x})$, and $\mathbf{M}_{\tilde{\mathbf{x}}, \mathbf{y}'}=\mathcal{G}_\phi\left(\mathcal{E}_\theta(\tx), \y'\right)$.
    
\subsection{Target Attribution Network}
    The TAN provides adequate assistance to the CMG by incorporating a target label to the synthesized conditioned map. Specifically, the objective of the TAN is to guide the generator to produce counterfactual maps that transform an input sample to be classified as a target class:
    \begin{equation}
            \mathcal{L}_{cls} = \mathbb{E}_{\mathbf{x}\sim P_{\mathbf{x}}, \mathbf{y}\sim P_{\mathbf{y}}}[CE(\mathbf{y}, \Tilde{\mathbf{y}})],
    \end{equation}
    where $\y$ denote the random target label, $\tilde{\y}=\mathcal{F}(\tilde{x})$, and $CE$ denote the cross-entropy function.
    
     Conceptually, the role of the TAN is similar to that of a discriminator in GANs, but their objective is very different. While a discriminator learns to distinguish between real and fake samples, the TAN is already trained in classifying the input samples. Thus, the discriminator plays a min-max game with a generator in an effort to produce more realistic samples, while the TAN provides deterministic guidance for the generator to produce class-specific samples.
    
\subsubsection{Counterfactual Map Loss}
    The counterfactual map loss limits the values of the counterfactual map to grow:
    \begin{equation}
            \mathcal{L}_{map} = \epx[\lambda_1\|\mathbf{M}_{\mathbf{x}, \mathbf{y}}\|_1 + \lambda_2\|\mathbf{M}_{\mathbf{x}, \mathbf{y}}\|_2],\label{cm loss}
    \end{equation}
    where $\lambda_1$ and $\lambda_2$ are weighting constants as the hyperparameter, and $\mathbf{y}$ is the random target label. This loss is crucial in solving two issues in counterfactual map generation. First, when left untethered, the counterfactual map will destroy the identity of the input sample. This problem is related to adversarial attacks~\cite{xu2018structured}, where a simple perturbation to the input sample will change the model's decision. Second, from an end-user point of view, only the most important features should be represented as the counterfactual explanation. Thus, by placing a constraint on the magnitude of counterfactual maps, we can address the above issues in one step.

\begin{figure}[t]
    \centering
    {\includegraphics[width=.9\linewidth]{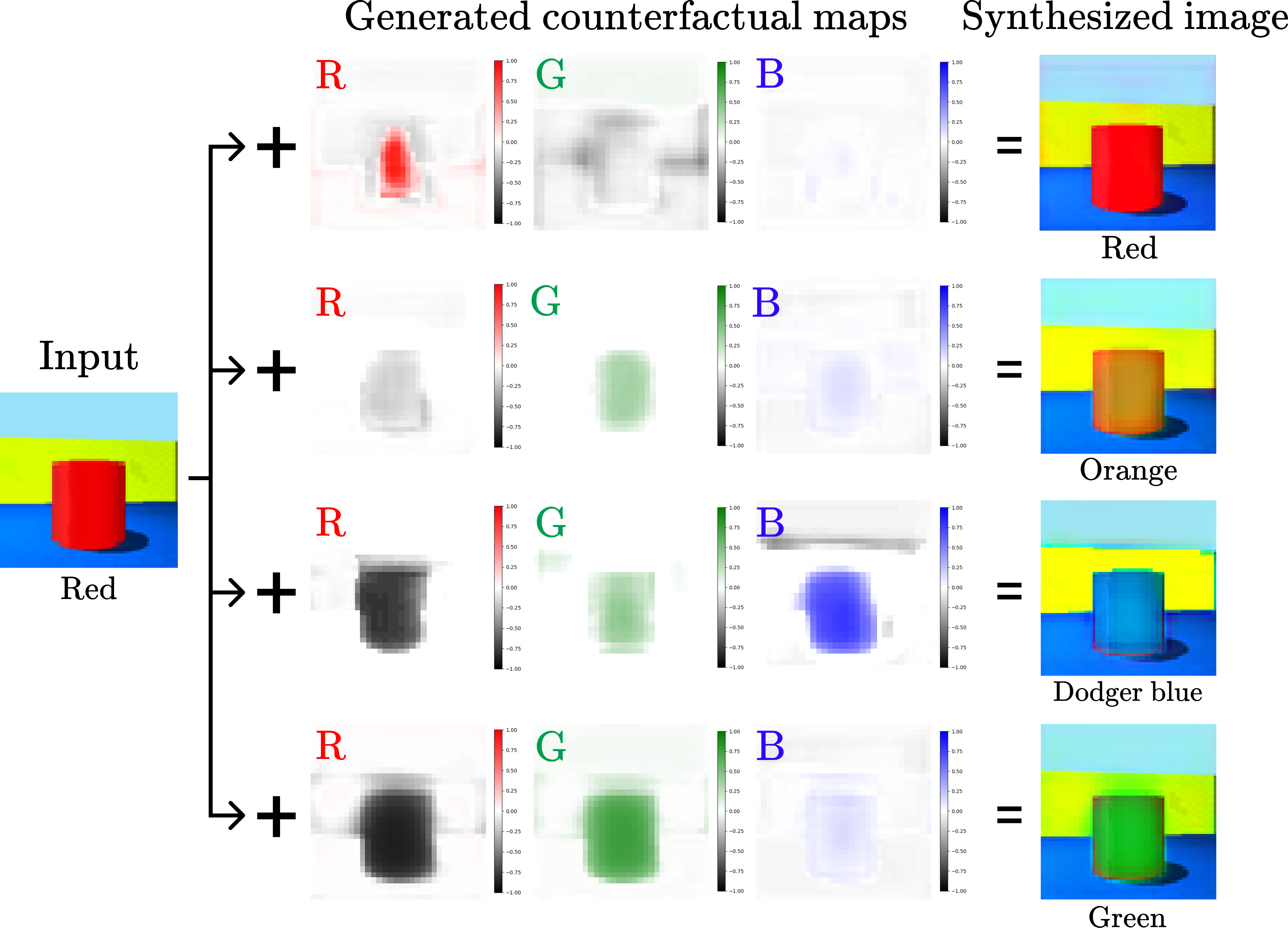}
    \caption{Examples of counterfactual maps for 3D Shapes~\cite{3dshapes18} dataset. The counterfactual map consists of the RGB channel. This is the outcome of producing a counterfactual map where the red cylinder is classified as red, orange, dodger blue, and green hue cylinders.}\label{fig:geometric_exp}}
\end{figure}

\subsection{Learning}
        Finally, we define the overall loss function for BIN as follows:
        \begin{equation}
            \mathcal{L}=\lambda_{3}\mathcal{L}^{\mathcal{D}_\psi}_{adv}+\lambda_{4}\mathcal{L}^{\mathcal{G}_\phi}_{adv}+\lambda_{5}\mathcal{L}_{cyc}+\lambda_{6}\mathcal{L}_{cls}+\mathcal{L}_{map},
        \end{equation}
        where $\lambda$ is the hyperparameter of the model ($\lambda_1$ and $\lambda_2$ in Eq.~\ref{cm loss}).
        
        During training, we share and fix the weights of the encoder $\mathcal{E}_\theta$ of the CMG with the TAN $\mathcal{F}$ to ensure that the attribution is consistent throughout the generative process. However, preliminary experiments on training the encoder $\mathcal{E}_\theta$ from scratch resulted in slightly lower qualitative and quantitative outcomes.

\section{Experiments}
\label{subsec:experiments}
    We conduct various experiments to validate the counterfactual maps generated by our proposed method. First, we conducted a qualitative analysis of a multi-way counterfactual explanation. Second, we reported a quantitative analysis using a correlation measure between generated counterfactual maps and ground-truth maps. Finally, we performed a suite of ablation studies to verify that each component of the BIN towards creating a meaningful counterfactual map.

\begin{figure}[t]
    \centering
    {\includegraphics[width=.95\linewidth]{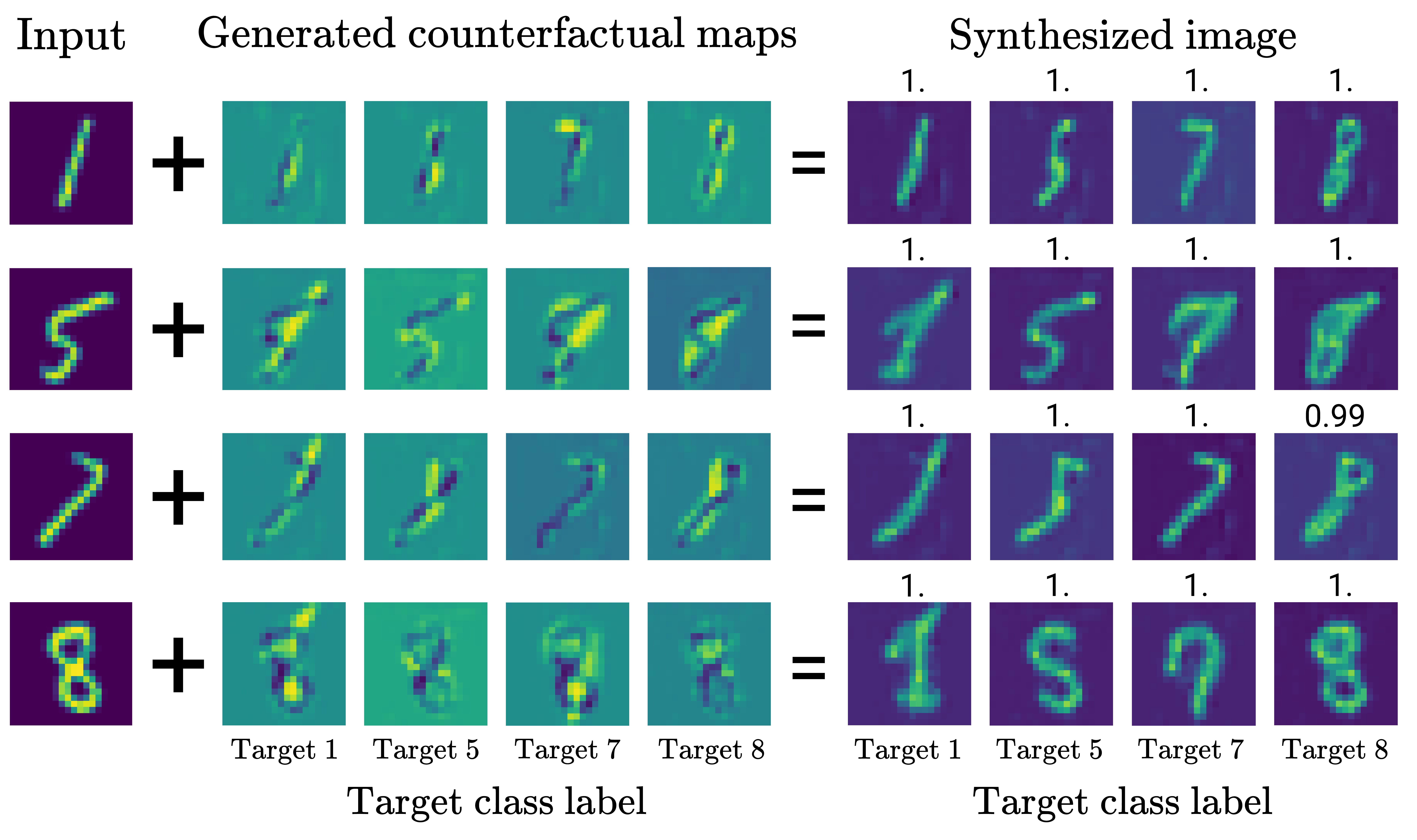}
    \caption{Examples of counterfactual maps for MNIST dataset. The resulting synthesized image is a addition between an input and its corresponding counterfactual map conditioned on a target label. The values on the top of each synthesized map is the model's softmax activated logit.}\label{fig:MNIST}}
\end{figure}

\begin{figure}[t]
    \centering
    {\includegraphics[width=.95\columnwidth]{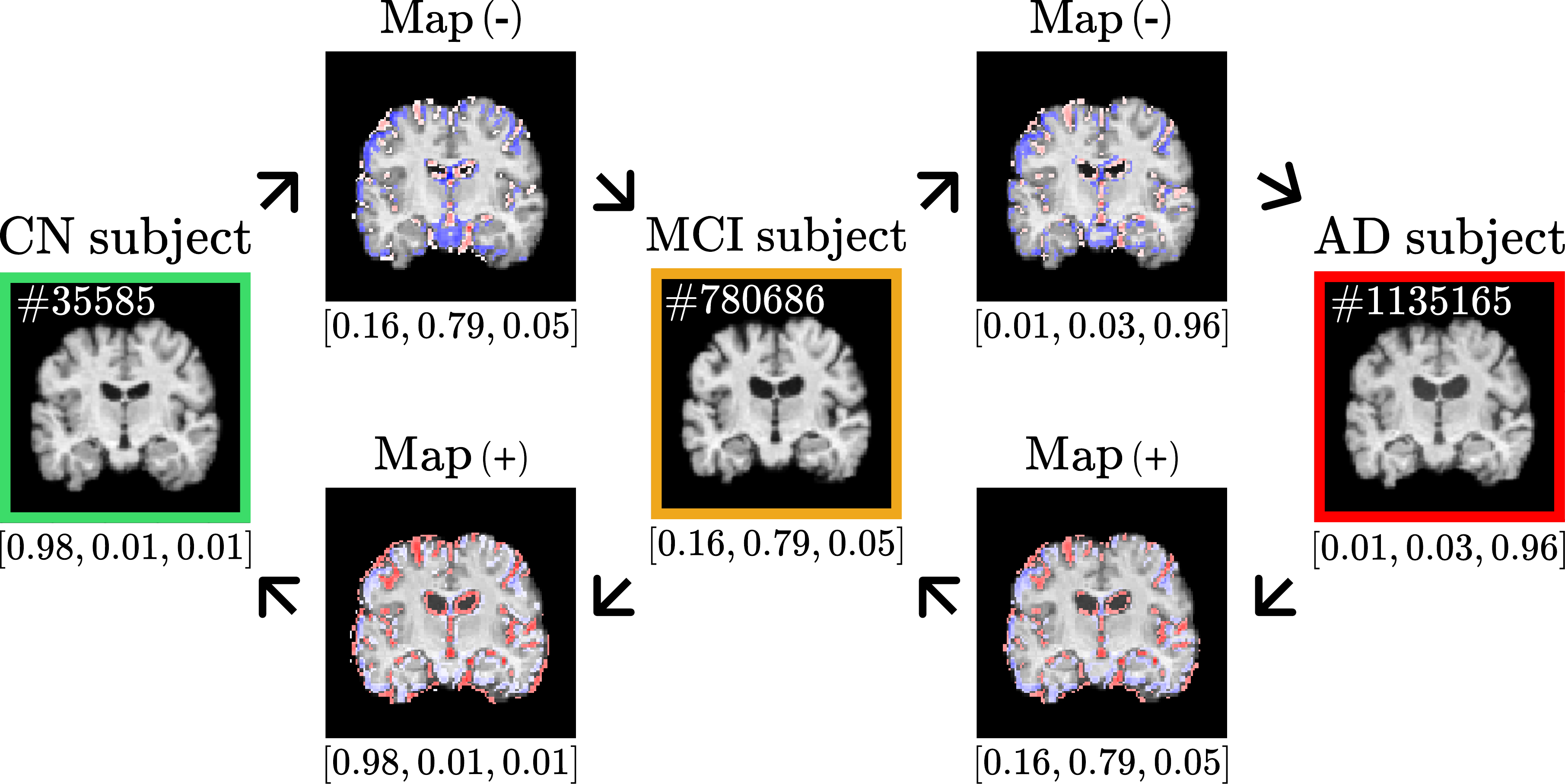}
    \caption{Examples of counterfactual maps for ADNI dataset (Subject ID 023\_S\_1190, Image ID on top left corner). Boundary boxes for green, orange, and red indicate ground-truths corresponding to CN, MCI, and AD, respectively, and the prediction scores at the bottom of each image are the model's softmax activated logit.  }\label{fig:multi_interpolation}}
\end{figure}

\subsection{Datasets}
    % For the comprehensive analysis of our work, we have chosen three data domains to perform our experiments: 3D Shapes~\cite{3dshapes18}, MNIST~\cite{lecun1998mnist}, and ADNI (Alzheimer's Disease Neuroimaging Initiative~\cite{MUELLER2005869}.
    
    % For a comprehensive analysis of our work, we have chosen three data domains to perform our experiments. First, 3D Shapes dataset~\cite{3dshapes18} is used for comparing dual-way counterfactual explanations. Second, MNIST~\cite{lecun1998mnist} is used for multi-way counterfactual explanations. Finally, ADNI (Alzheimer's Disease Neuroimaging Initiative)~\cite{MUELLER2005869} dataset is used as a real-world application in the medical domain.
    
\begin{figure*}[h]
    \centering
    {\includegraphics[width=0.9\linewidth]{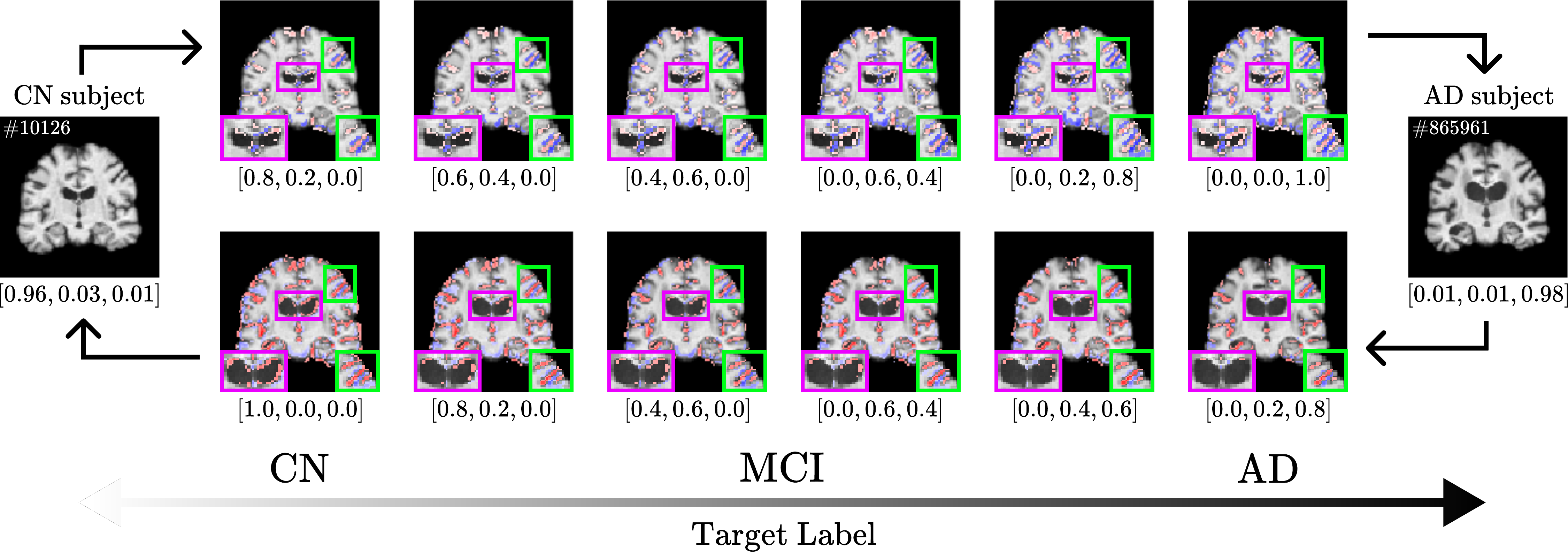}
    \caption{Example of counterfactual map conditioned on interpolated target labels (Subject ID 123\_S\_0106, Image ID on top left corner). The purple and green boxes correspond to the ventricular and cortex regions, respectively.
    The values at the bottom of left-most and right-most images are the model's softmax activated logits, and the values at the bottom of top and bottom rows (\ie, interpolated images) are the target labels.}\label{fig:brain_interpolation}}
\end{figure*}

\paragraph{3D Shapes}
    The 3D Shapes dataset~\cite{3dshapes18} consists of 480,000 RGB images of 3D geometric shapes with six latent factors (10 floors/wall/object hues, 8 scales, 4 shapes, and 15 orientations). For our experiments, we selected red, orange, dodger blue, and green object hues as the target classes. We randomly divided the dataset into a train, validation, and test set at a ratio of 8:1:1 and applied channel-wise min-max normalization.

\paragraph{MNIST} For MNIST dataset, we used the data split provided by~\cite{lecun1998mnist} and applied min-max normalization.

\paragraph{ADNI}
    ADNI dataset~\cite{MUELLER2005869} consists of 3D Magnetic Resonance Imaging (MRI) of various subject groups ranging from Cognitive Normal (CN) to Alzheimer's Disease (AD). For this dataset, we have selected a baseline MRI of 433 CN subjects, 748 Mild Cognitive Impairment (MCI) subjects, and 359 AD subjects in ADNI 1/2/3/Go studies. MCI is considered as a prodromal stage (\ie, preceding stage) of AD~\cite{silveira2010boosting}, which is a transitional state between normal cognitive changes and early clinical symptoms of dementia.
    
    For longitudinal studies, we have selected 12 CN test subjects that have converted to the AD group after the MCI stage at any given time. With the exception of 12 longitudinal test subjects, all subjects were randomly split into train, validation, and test sets at a ratio of 8:1:1. The pre-processing procedure consisted of neck removal (FSL v6.0.1 robustfov), brain extraction (HDBet~\cite{https://doi.org/10.1002/hbm.24750}), linear registration (FSL v6.0.1 FLIRT), zero-mean unit-variance normalization, quantile normalization at 10\% and 90\%, and down-scaling by $2\times$. The resulting pre-processed MRI is a $96 \times 114 \times 96$ image. We used default parameters from FSL v6.0.1~\cite{JENKINSON2012782}.

\subsection{Implementation}
    For 3D Shapes and MNIST experiments, we re-implemented the model from Kim \etal.~\cite{kim2018disentangling} with minor modifications. SonoNet-16~\cite{baumgartner2017sononet} is used for ADNI dataset as the encoder $\mathcal{E}_\theta$. The generator $\mathcal{G}_\phi$ has the same network design as the encoder $\mathcal{E}_\theta$ with pooling layers replaced by up-sampling layers. The test accuracy of the pre-trained TAN $\mathcal{F}$ is 99.56\% for MNIST, 99.81\% for 3D Shapes, and 73.79\% for ADNI dataset, respectively. 
    For compared methods, \ie, attribution-based approaches (Integrated gradients~\cite{sundararajan2017axiomatic}, LRP-Z~\cite{bach2015pixel}, DeepLift~\cite{shrikumar2017learning}) and perturbation-based approach (Guided backpropagation~\cite{selvaraju2017grad}), we used the pre-trained TAN $\mathcal{F}$ as the classifier.
    Further implementation details are in the Supplementary.

\subsection{Qualitative Analysis}                                                  
    An \textit{N-way} counterfactual map is a map that when added to an input sample, the classifier classifies it as a targeted class. The \textit{N-way} term refers to the degree of freedom of the targeted class. %For example, a multi-way counterfactual map can transform a red cylinder to be classified as red, orange, dodger blue, and green hues (Figure \ref{fig:geometric_exp}).
    In the following sections, we compared and validated multi-way counterfactual maps in various settings. Specifically, we took a bottom-up analysis approach with experiments on a toy example building up to application in a real-world dataset. First, we analyzed our novel multi-way counterfactual maps using an easy-to-understand toy example, which is the 3D Shapes and MNIST dataset. Additionally, we applied our proposed BIN in a real-world dataset and compared it with state-of-the-art XAI frameworks with ADNI dataset.

\begin{figure*}[h]
    \centering
    {\includegraphics[width=0.85\linewidth]{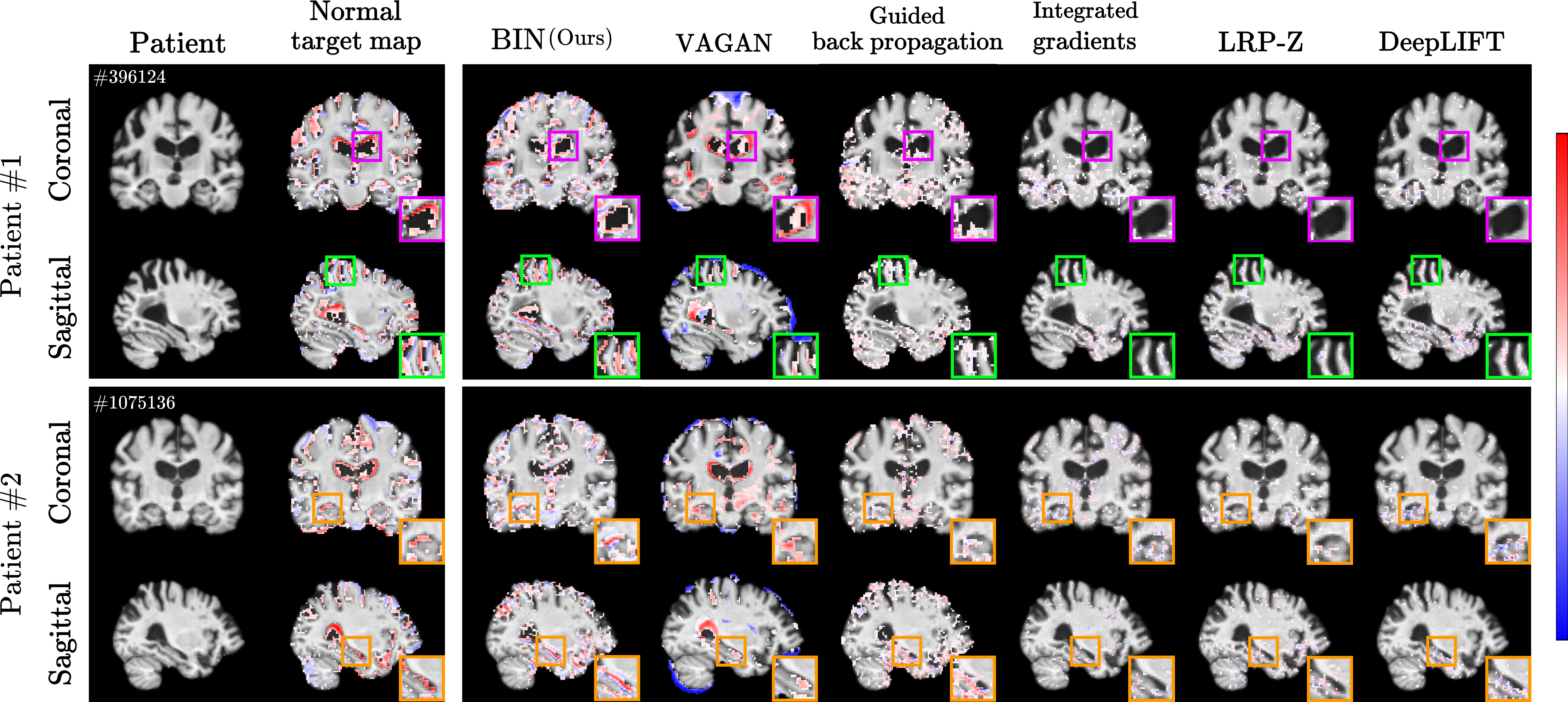}
    \caption{Example of counterfactual maps for ADNI dataset (Patient \#1 ID 024\_S\_0985, patient \#2 ID 002\_S\_4225, Image ID on top left corner). Purple, green, orange boxes visualize ventricular, cortex, and hippocampal regions, respectively.}\label{fig:Brain_result}}
\end{figure*}

\subsubsection{Validation of Counterfactual Map}
    \paragraph{3D Shapes}
    We analyzed multi-way counterfactual maps using 3D Shapes dataset (Figure \ref{fig:geometric_exp}). The goal of this experiment is to verify that our counterfactual map is able to transform an input image with regard to a target latent factor, \ie, object hue. The first row shows the counterfactual map from red (R: 255, G: 0, B: 0) to red (R: 255, G: 0, B: 0), \ie, for transforming to its own hue. Although an ideal counterfactual map (\ie, zero-value map) was not generated, each color channel of the map is a minor variation that does not destroy the input's hue. %The second row shows a counterfactual map for transforming red (R: 255, G: 0, B: 0) to orange (R: 255, G: 165, B: 0). Transforming red to orange is a relatively simpler task since it only requires a transformation in the G channel.
    The third and fourth row transformed red (R:255, G: 0, B: 0) to dodger blue (R: 30, G: 144, B: 255) and green (R: 0, G: 255, B: 0), respectively. Despite the contrasting hue in all channels, the BIN was able to induce the counterfactual map corresponding to the target object hue. In addition, the classification performance on the synthesized image is 99.12\%, indicating that the counterfactual map has successfully transformed the input image to be classified as another label. Furthermore, we have performed an interpolation, and this result is included in the Supplementary.

    \paragraph{Invariance of Irrelevant Factors}
    In most counterfactual maps generated in this experiment, we have observed an interesting phenomenon in which the counterfactual maps are relatively \textit{invariant} to latent factors that were not targeted for transformation. For example, the scale and shape of the object in the counterfactual map are relatively similar to that of the input image. This indicates that our counterfactual map is somewhat localized to the target latent factor.

    \paragraph{MNIST}
    We generated multi-way counterfactual maps for MNIST dataset (Figure~\ref{fig:MNIST}). In plain sight, we can observe that the style of an input image is maintained. This was related to the invariant to non-targeted latent factors observed in 3D Shapes experiment. We hypothesize that this invariant is due to the counterfactual map loss (Eq.~(\ref{cm loss})) since it puts constraints on the values of the counterfactual map so that it transforms an input sample with the least amount of energy. From a conceptual view, our generated counterfactual map exhibited a good example of counterfactual reasoning since it can successfully produce hypothetical realities with regards to a given input sample.

\subsubsection{Application in Medical Domain}\label{ADNC_part}
    To validate that our BIN can be generalized in the medical domain, we investigated multi-way counterfactual maps using ADNI dataset (Figure \ref{fig:multi_interpolation}), which possesses suitable scenarios in terms of its progression stage.
    
    \paragraph{Multi-way Counterfactual Map Interpolation} One of the biomarkers of Alzheimer's disease is an atrophy in brain regions, such as the ventricular~\cite{de2006longitudinal}, cortical thickness~\cite{chetelat2002mapping,fan2008spatial}, and hippocampus~\cite{iqbal2005subgroups,lindberg2012hippocampal,gerardin2009multidimensional}.
    As observed in the multi-way counterfactual maps (Figure~\ref{fig:brain_interpolation}), atrophy is clearly visible in those regions. 
    % {\modification VAGAN generated that the counterfactual map of regions similar to the highlighted atrophy in the Normal target map. Even though the aforementioned VAGAN was optimized for counterfactual map generation, our proposed BIN clearly captured neuropathological changes throughout the brain. That is, the BIN can observe subtle changes (\ie, potential biomarker regions) even in 3D brain images with high complexity.}
    
    \paragraph{Normal Target Map} Since there are no ground-truth maps for ADNI dataset, we utilized the Normal target map from longitudinal test subjects (second column in Figure~\ref{fig:Brain_result}).
    First, we have gathered MRIs from 12 subjects who converted from CN (baseline) to MCI and AD at any given time (subject ID in the Supplementary). Then, to create the Normal target map, we subtracted the baseline image from the target class image. This Normal target map exhibited a good representation of a ground-truth map of disease localization since we can observe which regions are possibly responsible for the conversion. The counterfactual map generated by attribution-based approaches (Integrated gradients, LRP-Z, DeepLift) and perturbation-based approach (Guided backpropagation) does not clearly showed the regions responsible for.

\begin{table}[t]\scriptsize
    \caption{Normalized Cross-Correlation (NCC) scores for 3D Shapes and ADNI dataset$^2$.}
    \label{table:compared exp}
    \centering
    \begin{tabular}{ccccc}
    \toprule
    \multicolumn{1}{c}{\multirow{2}{*}{\textbf{Method}}} & \multicolumn{2}{c}{\textbf{3D Shapes}} & \multicolumn{2}{c}{\textbf{ADNI}}\\
    \cmidrule(lr){2-3} \cmidrule(lr){4-5}
    & \textbf{NCC(+)} & \textbf{NCC(-)} & \textbf{NCC(+)} & \textbf{NCC(-)} \\

\midrule
    LRP-Z~\cite{bach2015pixel} & \multicolumn{1}{c}{0.008} & \multicolumn{1}{c}{0.086} &\multicolumn{1}{c}{0.008} &\multicolumn{1}{c}{0.005} \\
    Integrated Gradients~\cite{sundararajan2017axiomatic} &\multicolumn{1}{c}{0.006} &\multicolumn{1}{c}{0.152} &\multicolumn{1}{c}{0.006} &\multicolumn{1}{c}{0.005}\\
    DeepLIFT~\cite{shrikumar2017learning} &\multicolumn{1}{c}{0.007} &\multicolumn{1}{c}{0.183} &\multicolumn{1}{c}{0.005} &\multicolumn{1}{c}{0.004}\\
    Guided Backprop~\cite{selvaraju2017grad} &\multicolumn{1}{c}{0.183} &\multicolumn{1}{c}{0.123} &\multicolumn{1}{c}{0.239} &\multicolumn{1}{c}{0.204}\\
    VAGAN~\cite{baumgartner2018visual} &\multicolumn{1}{c}{0.381} &\multicolumn{1}{c}{-} &\multicolumn{1}{c}{0.317} &\multicolumn{1}{c}{-}\\
    \midrule
    \textbf{BIN (ours)} &\multicolumn{1}{c}{\bf{0.537}} &\multicolumn{1}{c}{\bf{0.452}} &\multicolumn{1}{c}{\bf{0.364}} &\multicolumn{1}{c}{\bf{0.201}}\\
\bottomrule
\end{tabular}
\end{table}

\subsection{Quantitative Analysis}
    \label{quan_anal}
    In this section, we quantitatively evaluated our proposed BIN and compared it with the outcome of other methods. To quantitatively assess the quality of our generated counterfactual maps, we've calculated the Normalized Cross-Correlation (NCC) score between generated maps and ground-truth maps. NCC score measures the similarity between two samples in a normalized setting, \ie, Higher NCC scores denote higher similarity. Thus, NCC can be helpful when two samples have a different magnitude of signals. However, since MNIST dataset does not have ground-truth maps, we have performed a different quantitative evaluation in section~\ref{FID_part}. For the ADNI dataset, we used the Normal target map described in section~\ref{ADNC_part} as the ground-truth map. Detailed results for all multi-way cases (\eg, transform AD to MCI or CN to AD, etc.) are in the Supplementary.\footnotetext[2]{Results for VAGAN NCC(-) (0.397 for 3D Shapes and 0.298 for ADNI) is excluded for fair comparison since it is a single-way map which require separate models for NCC(+) and NCC(-).}
    %NCC(+) refers to the counterfactual map for transforming the red cylinder to the other cylinder for 3D Shapes, and {\think AD to MCI} for ADNI dataset (vice versa for NCC(-)).
    
    % In Table~\ref{table:compared exp}, we reported NCC scores for 3D Shapes and ADNI dataset. Attribution-based methods (LRP-Z, Integrated gradients, and DeepLift) tend to have lower scores since lower discriminative features can be ignored in domains with high complexity, which is evident in Figure~\ref{fig:Brain_result}. Counterfactual maps for VAGAN can only transform input from AD to MCI, thus NCC(-) scores cannot be calculated.
    % Our proposed BIN had a outstanding NCC(+) scores indicating that the addition operation is a stronger suit than the subtraction operation. A further discussion on why NCC(-) is lower in section~\ref{low ncc}.

\subsection{Ablation Studies}
    We conducted a suite of ablation studies to assess each component of the BIN in creating a counterfactual map. Specifically, we focused on ablation studies on the conditioned generator (denoted as $\y$), the Target Attribution Network loss ($\mathcal{L}_{cls}$), cycle consistency loss ($\mathcal{L}_{cyc}$), and the Counterfactual Map loss ($\mathcal{L}_{map}$).

\paragraph{Normalized Cross Correlation}
For the ADNI and 3D Shapes dataset, we calculated NCC scores with the same settings as section~\ref{quan_anal} (Table ~\ref{table:Ablation studies}). The reported scores indicate every component of the BIN was crucial for producing a meaningful counterfactual map.
However, ablating the TAN loss $\mathcal{L}_{cls}$ showed a significant drop in NCC scores, indicating that it is one of the most crucial components of the BIN. The TAN guided the generative process to build targeted class attribution, which works in a similar manner to a discriminator in GANs. The cycle consistency loss $\mathcal{L}_{cyc}$ ensured the counterfactual map to contain information on its identity, \ie, input sample. In a preliminary experiment, we have observed that cycle consistency loss helps generate more crisp counterfactual maps, indicating that it works as a regularizer for the BIN. The counterfactual map loss $\mathcal{L}_{map}$ was another important component of the BIN since it regulates which region is most important.

\begin{table}[t]\scriptsize \setlength{\tabcolsep}{9.2pt}
    \caption{Normalized Cross-Correlation (NCC) scores for ablation studies.} 
    \label{table:Ablation studies}
    \centering
    \begin{tabular}{ccccc}
    \toprule
    \multicolumn{1}{c}{\textbf{Removed}} & \multicolumn{2}{c}{\textbf{3D Shapes}} & \multicolumn{2}{c}{\textbf{ADNI}}\\
    \cmidrule(lr){2-3} \cmidrule(lr){4-5}
    \textbf{Components}& \textbf{NCC(+)} & \textbf{NCC(-)} & \textbf{NCC(+)} & \textbf{NCC(-)} \\

\midrule
    $\y$ & \multicolumn{1}{c}{0.290} & \multicolumn{1}{c}{0.225} &\multicolumn{1}{c}{0.234} &\multicolumn{1}{c}{0.112} \\
    $\mathcal{L}_{cls}$ &\multicolumn{1}{c}{0.213} &\multicolumn{1}{c}{0.145} &\multicolumn{1}{c}{0.066} &\multicolumn{1}{c}{0.073}\\
    $\mathcal{L}_{cyc}$ &\multicolumn{1}{c}{0.501} &\multicolumn{1}{c}{0.362} &\multicolumn{1}{c}{0.279} &\multicolumn{1}{c}{0.145}\\
    $\mathcal{L}_{map}$ &\multicolumn{1}{c}{0.311} &\multicolumn{1}{c}{0.212} &\multicolumn{1}{c}{0.267} &\multicolumn{1}{c}{0.160}\\
    $\mathcal{L}_{cls},\mathcal{L}_{map}$ &\multicolumn{1}{c}{0.096} &\multicolumn{1}{c}{0.088} &\multicolumn{1}{c}{0.072} &\multicolumn{1}{c}{0.080}\\
    All above &\multicolumn{1}{c}{0.029} &\multicolumn{1}{c}{0.070} &\multicolumn{1}{c}{0.062} &\multicolumn{1}{c}{0.051}\\ 
    \midrule
    \textbf{BIN (ours)} &\multicolumn{1}{c}{\bf{0.537}} &\multicolumn{1}{c}{\bf{0.452}} &\multicolumn{1}{c}{\bf{0.364}} &\multicolumn{1}{c}{\bf{0.201}}\\ 
\bottomrule
\end{tabular}
\end{table}

\paragraph{Fréchet Inception Distance}
\label{FID_part}
For a quantitative assessment of multi-way counterfactual map, we used a Fréchet Inception Distance~\cite{heusel2017gans} commonly used for assessing the generative performance of GANs. Here, we've selected 4,000 test samples for counterfactual map generation (\ie fake images), and 4,000 test samples as real images. Since a counterfactual map can transform any number to any other number, a single image can generate 10 images (one per number). Thus, the total number of fake samples we've generated is 40,000 (\ie, 4,000 fake images per number). We compared our work in an ablation study setting (table in the Supplementary). In most of the settings, our proposed BIN performed significantly better.

\section{Discussion}
    \label{low ncc}
    For most experiments, reported NCC(-) scores for almost every compared method were lower than that of NCC(+). A possible explanation is that the activation functions, such as ReLU, may be imposing a positive bias or negative skewness on the model output. For example, the mean and mode of a counterfactual map for almost every datasets and every method compared in this paper were slightly over zero. This may result in imposing a constraint for subtraction operations which lowers the NCC(-) scores. Possible future work for the community is to verify whether this bias or skewness really exists and propose a way to alleviate this.

\section{Conclusion}
    In this work, we proposed Born Identity Network (BIN), which is a post-hoc approach for producing multi-way counterfactual maps in multiple hypothetical scenarios. By visualizing counterfactual explanations at the level of human knowledge through our BIN, classifiers that have derived high performance in various domains can provide explainability and interpretability of higher quality. It is expected that BIN will be a practical application that helps diagnose or predict disease through a counterfactual visual explanation of its applicability in the multi-classifier dealing with the medical field.

{\small
\bibliographystyle{ieee_fullname}
\bibliography{main}
}

\newpage
\onecolumn

\input{supplementary.tex}

\end{document}

%% file: supplementary.tex
\renewcommand\thefigure{S\arabic{figure}}
\renewcommand\thetable{S\arabic{table}}
\setcounter{figure}{0}
\setcounter{table}{0}
\renewcommand\thesection{\thechapter \Alph{section}} 
\renewcommand\thesubsection{\Alph{section}.\arabic{subsection}}

%%%%%%%%%%%%%%%%%%%%%%%%%%%%%%%%%%%%%%%%%%%%%%%%%%%%%%%
\setcounter{page}{11} %11

\begin{center}\LARGE\bfseries
Supplementary
\end{center}

\section*{Section 1: Extra Counterfactual Map Interpolation}
\label{3Dshape_interpolation}

\begin{figure}[h]
    \centering
	\includegraphics[width=0.85\linewidth]{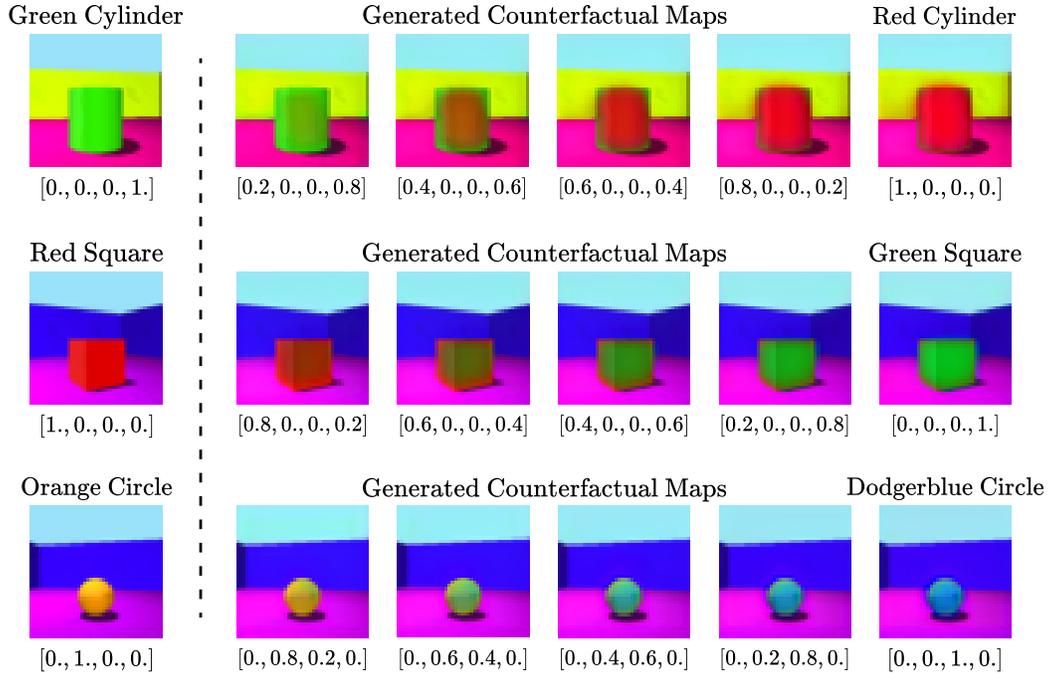}
	\caption{Example of counterfactual map for 3D Shapes dataset conditioned on interpolated target labels. The one-hot vector at the bottom of each counterfactual map is the target labels.}
\end{figure}
\begin{figure}[h]
    \centering
	\includegraphics[width=0.86\linewidth]{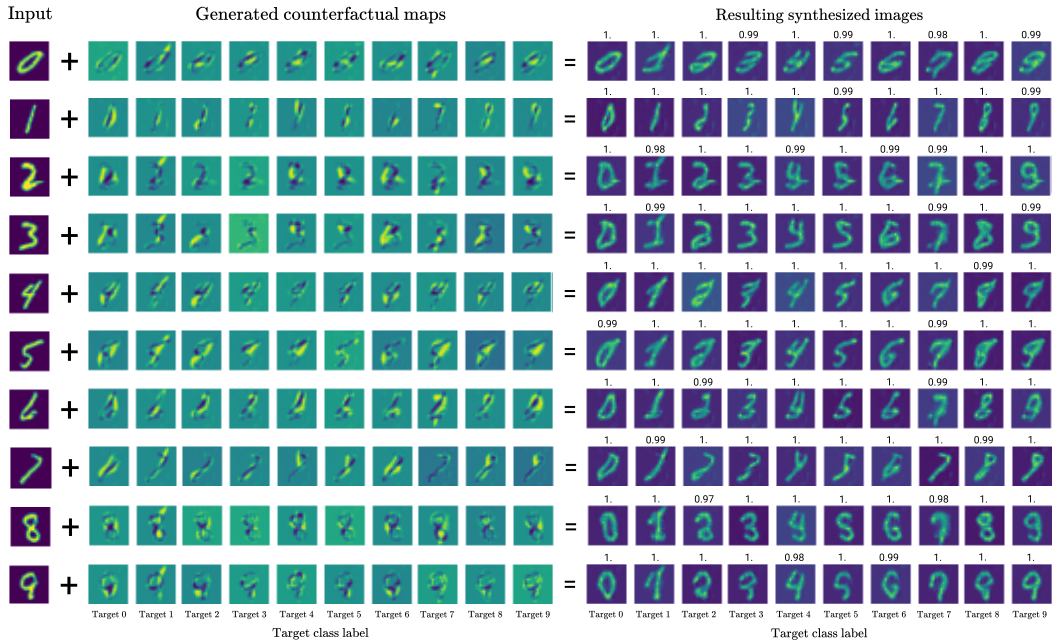}
	\caption{Examples of counterfactual maps for MNIST dataset. The resulting synthesized image is an addition between an input and its corresponding counterfactual map conditioned on a target label. The values on the top of each confound map are the model's softmax activated logit.}
\end{figure}

%%%%%%%%%%%%%%%%%%%%%%%%%%%%%%%%%%%%%%%%%%%%%%%%%%%%%%%

\section*{Section 2: NCC Scores in Multi-way Settings}
\label{multiway_quan}
For an in-depth analysis of the our \textit{multi-way} counterfactual maps, we calculated NCC(+) and NCC(-) scores in a multi-class setting (\ie, AD $\leftrightarrow$ CN, AD $\leftrightarrow$ MCI, and MCI $\leftrightarrow$ CN). With the exception of VAGAN, all methods utilize the same 3-class (CN \textit{vs.} MCI \textit{vs.} AD) classifier. Note that VAGAN is a single-way method with the assumption that the label of observation is known. In other words, it is not possible to obtain scores of NCC(+) and NCC(-) simultaneously from a single model. The results of VAGAN in Table~\ref{table:adni_multiway_ncc} are from six different models, while all other methods are from the same classifier.

\begin{table}[h]
    \renewcommand{\arraystretch}{1.}
    \caption{Normalized Cross-Correlation (NCC) scores for all multi-way cases in ADNI dataset.}
    \label{table:adni_multiway_ncc}
    \centering
    \begin{tabular}{ccccccc}
    \toprule
    \multicolumn{1}{c}{\multirow{2}{*}{\textbf{Method}}} & \multicolumn{2}{c}{\textbf{AD $\leftrightarrow$ CN}} & \multicolumn{2}{c}{\textbf{AD $\leftrightarrow$ MCI}} & \multicolumn{2}{c}{\textbf{MCI $\leftrightarrow$ CN}}\\
    \cmidrule(lr){2-3} \cmidrule(lr){4-5} \cmidrule(lr){6-7}
    & \textbf{NCC(+)} & \textbf{NCC(-)} & \textbf{NCC(+)} & \textbf{NCC(-)} & \textbf{NCC(+)} & \textbf{NCC(-)} \\

\midrule
    LRP-Z~\cite{bach2015pixel} &\multicolumn{1}{c}{0.008} &\multicolumn{1}{c}{0.005} &\multicolumn{1}{c}{0.006} &\multicolumn{1}{c}{0.004} &\multicolumn{1}{c}{0.005} &\multicolumn{1}{c}{0.005} \\
    Integrated Gradients~\cite{sundararajan2017axiomatic} &\multicolumn{1}{c}{0.006} &\multicolumn{1}{c}{0.005} &\multicolumn{1}{c}{0.007} &\multicolumn{1}{c}{0.007} &\multicolumn{1}{c}{0.006} &\multicolumn{1}{c}{0.007}\\
    DeepLIFT~\cite{shrikumar2017learning} &\multicolumn{1}{c}{0.005} &\multicolumn{1}{c}{0.004} &\multicolumn{1}{c}{0.006} &\multicolumn{1}{c}{0.004} &\multicolumn{1}{c}{0.004} &\multicolumn{1}{c}{0.005}\\
    Guided Backprop~\cite{selvaraju2017grad} &\multicolumn{1}{c}{0.239} &\multicolumn{1}{c}{0.204} &\multicolumn{1}{c}{0.212} &\multicolumn{1}{c}{0.163} &\multicolumn{1}{c}{0.199} &\multicolumn{1}{c}{0.158}\\
    VAGAN~\cite{baumgartner2018visual} &\multicolumn{1}{c}{0.317} &\multicolumn{1}{c}{\bf{0.298}} &\multicolumn{1}{c}{0.285} &\multicolumn{1}{c}{\bf{0.257}} &\multicolumn{1}{c}{0.283} &\multicolumn{1}{c}{0.186}\\
    \midrule
    \textbf{BIN (ours)} &\multicolumn{1}{c}{\bf{0.364}} &\multicolumn{1}{c}{0.201} &\multicolumn{1}{c}{\bf{0.301}} &\multicolumn{1}{c}{0.174} &\multicolumn{1}{c}{\bf{0.292}} &\multicolumn{1}{c}{\bf{0.188}}\\
\bottomrule
\end{tabular}
\end{table}

%%%%%%%%%%%%%%%%%%%%%%%%%%%%%%%%%%%%%%%%%%%%%%%%%%%%%%%

\section*{Section 3: Ablation Studies - Fréchet Inception Distance}
\label{MNIST_fid}
Since there are no ground truth counterfactual maps (\ie, Normal target maps) for MNIST dataset, we calculated the Fréchet Inception Distance~\cite{heusel2017gans} score, which is commonly used to assess the synthesized images of generative models. In addition, we performed a suite of ablations studies on core components of BIN.

\begin{table}[h]
\renewcommand{\arraystretch}{.9}
\centering
\caption{Fréchet Inception Distance (FID) scores reported for ablation studies for MNIST dataset.}
\label{table:Ablation studies fid}
\begin{tabular}{ccccccccccccccc}
\toprule
    \multicolumn{4}{c}{\bfseries Components} &\multicolumn{10}{c}{\bfseries FID score} \\ 
    \multicolumn{1}{c}{$\y$} &\multicolumn{1}{c}{$\mathcal{L}_{cls}$} &\multicolumn{1}{c}{$\mathcal{L}_{cyc}$} &\multicolumn{1}{c}{$\mathcal{L}_{map}$} &\textbf{0} &\textbf{1} &\textbf{2} &\textbf{3} &\textbf{4} &\textbf{5} &\textbf{6} &\textbf{7} &\textbf{8} &\textbf{9} &\textbf{avg}\\ 
    \cmidrule(lr){1-4} \cmidrule(lr){5-15}
    
    \multicolumn{1}{c}{} &\multicolumn{1}{c}{\checkmark} &\multicolumn{1}{c}{\checkmark} &\multicolumn{1}{c}{\checkmark} & \text{1.340} &\text{1.082} &\text{1.034} &\text{0.935} &\text{0.945} &\text{0.933} &\text{0.999} &\text{0.850} &\text{0.961} &\text{0.850} &\text{0.993}\\
    
    \multicolumn{1}{c}{\checkmark} &\multicolumn{1}{c}{} &\multicolumn{1}{c}{\checkmark} &\multicolumn{1}{c}{\checkmark} & \text{1.338} &\text{1.241} &\text{1.039} &\text{0.955} &\text{1.064} &\text{0.973} &\text{1.022} &\text{0.956} &\text{0.993} &\text{0.935} &\text{1.052} \\
    
    \multicolumn{1}{c}{\checkmark} &\multicolumn{1}{c}{\checkmark} &\multicolumn{1}{c}{} &\multicolumn{1}{c}{\checkmark} & \text{1.283} &\text{1.156} &\text{0.915} &\text{0.810} &\text{0.959} &\text{0.822} &\text{0.987} &\text{0.888} &\textbf{0.896} &\text{0.834} &\text{0.955} \\
    
    \multicolumn{1}{c}{\checkmark} &\multicolumn{1}{c}{\checkmark} &\multicolumn{1}{c}{\checkmark} &\multicolumn{1}{c}{} & \text{1.318} &\text{1.056} &\text{0.906} &\text{0.804} &\textbf{0.951} &\text{0.824} &\text{1.001} &\text{0.841} &\text{0.904} &\text{0.841} &\text{0.945} \\
    
    \multicolumn{1}{c}{\checkmark} &\multicolumn{1}{c}{} &\multicolumn{1}{c}{} &\multicolumn{1}{c}{\checkmark} & \text{1.338} &\text{1.242} &\text{1.039} &\text{0.956} &\text{1.065} &\text{0.973} &\text{1.022} &\text{0.956} &\text{0.993} &\text{0.935} &\text{1.052}\\
    
    \multicolumn{1}{c}{} &\multicolumn{1}{c}{} &\multicolumn{1}{c}{} &\multicolumn{1}{c}{} & \text{1.338} &\text{1.242} &\text{1.039} &\text{0.956} &\text{1.065} &\text{0.973} &\text{1.022} &\text{0.956} &\text{0.993} &\text{0.935} &\text{1.052}\\ \midrule
    
    \multicolumn{1}{c}{\checkmark} &\multicolumn{1}{c}{\checkmark} &\multicolumn{1}{c}{\checkmark} &\multicolumn{1}{c}{\checkmark} & \textbf{1.211} &\textbf{0.902} &\textbf{0.892} &\textbf{0.802} &\text{0.958} &\textbf{0.812} &\textbf{0.970} &\textbf{0.835} &\text{0.912} &\textbf{0.833} &\textbf{0.912}\\
\bottomrule
\end{tabular}
\end{table}

%%%%%%%%%%%%%%%%%%%%%%%%%%%%%%%%%%%%%%%%%%%%%%%%%%%%%%% 

\section*{Section 4: Preliminary Survey on Related Works}
\label{comparison_other_method}

\begin{table}[h]
\renewcommand{\arraystretch}{1.}
\centering
\caption{Preliminary Survey on Related Works}
\label{tab:survey}
\begin{tabular}{cccccc}
\toprule
\multicolumn{1}{c}{\multirow{2}{*}{\textbf{}}} & \multicolumn{1}{c}{\multirow{2}{*}{\textbf{Method}}} & \multicolumn{1}{c}{\multirow{2}{*}{\textbf{Way}}} & \multicolumn{1}{c}{\multirow{2}{*}{\textbf{Stratification}}} & \multicolumn{1}{c}{\multirow{2}{*}{\textbf{Counterfactual Map}}} & \multicolumn{1}{c}{\multirow{2}{*}{\textbf{Post-hoc}}}\\
&  &  &\\
\midrule
$\text{Sauer}$ $\etal.$~\cite{sauer2021counterfactual} & Generative & Single & $\times$ & $\surd$ & $\surd$ \\
$\text{Goyal}$ $\etal.$~\cite{goyal2019counterfactual} & Perturbation & Dual & $\times$ & $\times$ & $\surd$ \\
$\text{Baumgartner}$ $\etal.$~\cite{baumgartner2018visual} & Generative & Single & $\times$ & $\surd$ & $\times$ \\
$\text{Chang}$ $\etal.$~\cite{chang2018explaining} & Perturbation & Single & $\times$ & $\times$ & $\surd$ \\
$\text{Singla}$ $\etal.$~\cite{singla2019explanation} & Generative & Dual & $\surd$ & $\times$ & $\surd$ \\
$\text{Bass}$ $\etal.$~\cite{bass2020icam} & Generative & Dual & $\surd$ & $\surd$ & $\times$ \\ 
$\textbf{Ours}$ & Generative & Multi & $\surd$ & $\surd$ & $\surd$\\ 

% $\text{Sauer}$ $\etal.$~\cite{sauer2021counterfactual} & Generative & 1 &  & $\surd$ & $\surd$ \\
% $\text{Goyal}$ $\etal.$~\cite{goyal2019counterfactual} & Perturbation & 2 &  & & $\surd$ \\
% $\text{Baumgartner}$ $\etal.$~\cite{baumgartner2018visual} & Generative & 1 &  & $\surd$ & \\
% $\text{Chang}$ $\etal.$~\cite{chang2018explaining} & Perturbation & 1 &  & & $\surd$ \\
% $\text{Singla}$ $\etal.$~\cite{singla2019explanation} & Generative & 2 & $\surd$ & & $\surd$ \\
% $\text{Bass}$ $\etal.$~\cite{bass2020icam} & Generative & 2 & $\surd$ & $\surd$ &  \\ 
% $\textbf{Ours}$ & Generative & $\textit{N}$ & $\surd$ & $\surd$ & $\surd$\\ 

% $\text{Sauer}$ $\etal.$~\cite{sauer2021counterfactual} & Generation-based & $\surd$ & Global Explanation & Post-hoc \\
% $\text{Goyal}$ $\etal.$~\cite{goyal2019counterfactual} & Perturbation-based & $\times$ & Local Explanation & Post-hoc \\
% $\text{Baumgartner}$ $\etal.$~\cite{baumgartner2018visual} & Generation-based & $\surd$ & Global Explanation & Intrinsic\\
% $\text{Chang}$ $\etal.$~\cite{chang2018explaining} & Perturbation-based & $\times$ & Local Explanation & Post-hoc \\
% $\text{Singla}$ $\etal.$~\cite{singla2019explanation} & Generation-based & $\times$ & Local Explanation & Post-hoc \\
% $\text{Bass}$ $\etal.$~\cite{bass2020icam} & Generation-based & $\surd$ & Global Explanation & Intrinsic \\ 
% $\textbf{Ours}$ & Generation-based  & $\surd$ & Global Explanation & Post-hoc\\
\bottomrule
\end{tabular}
\end{table}
\clearpage
The generation-based method is an interpretation model that uses a generative process to produce visual explanations.
There are roughly two different manners in which generative models approach visual explanations: generative explanation, and counterfactual explanation. Generative explanation ($\ie,$ generating $\Tilde{\mathbf{x}}$) approach that focuses on generating images to be classified as a target class, rather than conducting analytic reasoning.
Counterfactual explanation approach ($\ie,$ generating $\mathbf{M}$, where $\Tilde{\mathbf{x}} = \mathbf{x} + \mathbf{M}$) questions $\textit{what}$ and $\textit{why}$ a sample was classified as a target class, which can be intuitively inferred through the generated counterfactual map.

Most works on counterfactual maps exploit single- or dual-way explanation, which can only provide one or two hypothetical alternatives for counterfactual reasoning.
Stratification, in the context of counterfactual reasoning, provides explanation in a continuous space.
This allows counterfactual maps to be generated for any point between two or more outcomes (\eg, normal and Alzheimer's disease), where without it, counterfactual maps can only explain extreme tails of the outcome.
However, stratified single-/dual-way counterfactual maps cannot model the complex and non-linear progression from one outcome (\eg, normal) to other (\eg, Alzheimer's disease).
Thus, they are only able to draw a linear interpolation between the two outcomes.
When stratification is combined with multi-way counterfactual maps, these generated maps can model a more precise and realistic progression from one outcome to the other.
Thus, multi-way stratified counterfactual maps allow explanation for predefined multiple outcomes, as well as interpolate to undefined outcomes.
% While the perturbation-based method is a feature attribution approach,

% which focus on the change in classifier outputs with respect to perturbed input images $\ie,$ input images 

% where parts of the image have been masked and replaced with various references such as mean pixel values, 

% blurred image areas, random noise, $\etc$. ?????

% Local explanation focuses on the specifics of each pixel or region and provides explanations that can lead to a better understanding of the contribution in smaller groups of certain features. Global explanation can provide a clearer and intuitive understanding than the local explanation in terms of model interpretation due to its ability to explain the importance of features in the whole area.

% Finally, the intrinsic manner is to build inherently interpretable modules or functions in models, and the post-hoc manner is building an interpretability method for already-trained models.

Non-post-hoc methods build interpretable models or modules from scratch, while post-hoc methods builds on top of a pretrained model. Thus, post-hoc methods are a more generalized interpretation method that can be applied to most connectionist models. 

%%%%%%%%%%%%%%%%%%%%%%%%%%%%%%%%%%%%%%%%%%%%%%%%%%%%%%%
\section*{Section 5: Details of Longitudinal Dataset}
\label{longitudinal_subjectID}
Since there are no ground-truth maps for ADNI dataset~\cite{MUELLER2005869}, we utilized the Normal target map from longitudinal test subjects. First, we have gathered MRIs from 12 subjects who converted from CN (baseline) to MCI and AD at any given time. Then, to create the Normal target map, we subtracted the baseline image from the target class image. This Normal target map exhibited a good representation of a ground-truth map of disease localization since we can observe which regions are possibly responsible for the conversion.

We have selected 12 CN test subjects that have converted to the AD group after the MCI stage at any given time. Details of the subject ID with regard to each subject and image ID corresponding to CN, MCI, and AD are shown in the Table~\ref{tab:adni_longitudinal_table}.

\begin{table}[h]
\renewcommand{\arraystretch}{1.}
\centering
\caption{ADNI longitudinal subject information for the quantitative evaluation of counterfactual map}
\label{tab:adni_longitudinal_table}
\begin{tabular}{cccc}
\toprule
\multicolumn{1}{c}{\multirow{2}{*}{\textbf{Subject ID}}} & \multicolumn{1}{c}{\textbf{Cognitive Normal}} & \multicolumn{1}{c}{\textbf{Mild Cognitive Impairment}} & \multicolumn{1}{c}{\textbf{Alzheimer's Disease}}\\
% \cmidrule(lr){3} \cmidrule(lr){4} \cmidrule(lr){5}
& \text{Image ID} & \text{Image ID} & \text{Image ID} \\
\cmidrule(lr){1-4}
$023\_\text{S}\_0061$ & 9046 & 401795 & 473765 \\
$123\_\text{S}\_0106$ & 10126 & 213947 & 865961 \\
$131\_\text{S}\_0123$ & 10042 & 292389 & 475755 \\
% $114\_\text{S}\_0166$ & 11018 & 289125 & 416112 \\
$005\_\text{S}\_0223$ & 11645 & 26115 & 143296 \\
$037\_\text{S}\_0467$ & 14861 & 169074 & 372375 \\
$129\_\text{S}\_0778$ & 20543 & 205611 & 388698 \\
$024\_\text{S}\_0985$ & 27607 & 342890 & 396124 \\
$023\_\text{S}\_1190$ & 35585 & 780686 & 1135165 \\
$002\_\text{S}\_4262$ & 259653 & 397601 & 788894 \\
$029\_\text{S}\_4385$ & 285589 & 421501 & 642470 \\
$098\_\text{S}\_4506$ & 286987 & 714545 & 1073489 \\
$051\_\text{S}\_5285$ & 396288 & 1055626 & 1182436 \\
\bottomrule
\end{tabular}
\end{table}

%%%%%%%%%%%%%%%%%%%%%%%%%%%%%%%%%%%%%%%%%%%%%%%%%%%%%%%%%%%%%%%%

\clearpage
\section*{Section 6: Network Architecture}
\label{network architecture}
% In each Table \ref{tab:shape_g}, \ref{tab:mnist_g}, and \ref{tab:adni_g}
$(\cdot)^*$ denotes the skip-connection layer that passed through the convolution layer of 3 kernel size, 1 stride after concatenation with the tiled target label. In the case of feature maps used in $(\cdot)^*$, we used the same number of feature maps of the counterfactual map generator to be concatenated.

We performed $n$-D convolutional operation in the order of Conv.$\rightarrow$BatchNorm./SpectralNorm.$\rightarrow$Activation. For the 3D Shapes and MNIST experiments, we re-implemented the encoder $\mathcal{E}_{\theta}$ from Kim \etal.~\cite{kim2018disentangling} with minor modification. In ADNI experiment, the encoder $\mathcal{E}_{\theta}$ network of SonoNet-16~\cite{baumgartner2017sononet} was used identically.

\subsection*{6-1. 3D Shapes}
\begin{table}[h]
\renewcommand{\arraystretch}{.95}
\centering
\caption{3D Shapes encoder $\mathcal{E}_{\theta}$ network.}
\label{tab:encoder}
\begin{tabular}{@{}rlcccccl@{}}
\toprule
Operation & Feature Maps & Batch Norm. & Kernels & Strides & Padding & Activation &\\ \midrule

Input $\mathbf{x}\in\mathbb{R}^{64 \times 64 \times 3}$ &  &  &  &  &  &  \\
(ENC1) 2D Conv. & \multicolumn{1}{c}{32} & $\surd$ & (4 $\times$ 4) & (2 $\times$ 2) & $\surd$ & \multicolumn{1}{c}{ReLU} \\
(ENC2) 2D Conv. & \multicolumn{1}{c}{32} & $\surd$ & (4 $\times$ 4) & (2 $\times$ 2) & $\surd$ & \multicolumn{1}{c}{ReLU} \\
(ENC3) 2D Conv. & \multicolumn{1}{c}{64} & $\surd$ & (4 $\times$ 4) & (2 $\times$ 2) & $\surd$ & \multicolumn{1}{c}{ReLU} \\
(ENC4) 2D Conv. & \multicolumn{1}{c}{64} & $\surd$ & (4 $\times$ 4) & (2 $\times$ 2) & $\surd$ & \multicolumn{1}{c}{ReLU} \\
Output $\mathbf{z}\in\mathbb{R}^{4 \times 4 \times 64}$ &  &  &  &  &  &  \\
\bottomrule
\end{tabular}
\end{table}

\begin{table}[h]
\renewcommand{\arraystretch}{.95}
\centering
\caption{3D Shapes classifier $\mathcal{F}$ network.}
\label{tab:encoder}
\begin{tabular}{@{}rlcccccl@{}}
\toprule
Operation & Feature Maps & Batch Norm. & Dropout & Activation &\\ \midrule
Input $\mathbf{z}\in\mathbb{R}^{4 \times 4 \times 64}$ &  &  &  &  &  &  \\
Fully Connected & \multicolumn{1}{c}{4} & $\times$ & $\times$ & \multicolumn{1}{c}{Softmax} \\

\bottomrule
\end{tabular}
\end{table}

\begin{table}[h]
\renewcommand{\arraystretch}{.95}
\centering
\caption{3D Shapes discriminator $\mathcal{D}_{\psi}$ network.}
\label{tab:encoder}
\begin{tabular}{@{}rlcccccl@{}}
\toprule
Operation & Feature Maps & Spectral Norm. & Kernels & Strides & Padding & Activation &\\ \midrule

Input $\mathbf{x}\in\mathbb{R}^{64 \times 64 \times 3}$ &  &  &  &  &  &  \\
2D Conv. & \multicolumn{1}{c}{32} & $\times$ & (4 $\times$ 4) & (2 $\times$ 2) & $\surd$ & \multicolumn{1}{c}{Leaky ReLU} \\
2D Conv. & \multicolumn{1}{c}{32} & $\surd$ & (4 $\times$ 4) & (2 $\times$ 2) & $\surd$ & \multicolumn{1}{c}{Leaky ReLU} \\
2D Conv. & \multicolumn{1}{c}{64} & $\surd$ & (4 $\times$ 4) & (2 $\times$ 2) & $\surd$ & \multicolumn{1}{c}{Leaky ReLU} \\
2D Conv. & \multicolumn{1}{c}{64} & $\surd$ & (4 $\times$ 4) & (2 $\times$ 2) & $\surd$ & \multicolumn{1}{c}{Leaky ReLU} \\
Fully Connected & \multicolumn{1}{c}{1} & $\times$ & & & & \multicolumn{1}{c}{Linear}\\
\bottomrule
\end{tabular}
\end{table}

\begin{table}[h]
\renewcommand{\arraystretch}{.95}
\centering
\caption{3D Shapes counterfactual map generator $\mathcal{G}_{\phi}$ network.}
\label{tab:shape_g}
\begin{tabular}{@{}rlcccccl@{}}
\toprule
Operation & Feature Maps & Batch Norm. & Kernels & Strides & Padding & Activation &\\ \midrule
Input $\mathbf{z}\in\mathbb{R}^{4 \times 4 \times 64}$ &  &  &  &  &  &  \\
(DEC3) Upsampling & & & & (2 $\times$ 2) & & \\
\multicolumn{5}{c}{Concatenate $\text{(ENC3)}^*$ and (DEC3) along the channel axis} \\
2D Conv. & \multicolumn{1}{c}{64} & $\surd$ & (3 $\times$ 3) & (1 $\times$ 1) & $\surd$ & \multicolumn{1}{c}{ReLU}\\
(DEC2) Upsampling & & & & (2 $\times$ 2) & & \\
\multicolumn{5}{c}{Concatenate $\text{(ENC2)}^*$ and (DEC2) along the channel axis} \\
2D Conv. & \multicolumn{1}{c}{32} & $\surd$ & (3 $\times$ 3) & (1 $\times$ 1) & $\surd$ & \multicolumn{1}{c}{ReLU}\\
(DEC1) Upsampling & & & & (2 $\times$ 2) & & \\
\multicolumn{5}{c}{Concatenate $\text{(ENC1)}^*$ and (DEC1) along the channel axis} \\
2D Conv. & \multicolumn{1}{c}{32} & $\surd$ & (3 $\times$ 3) & (1 $\times$ 1) & $\surd$ & \multicolumn{1}{c}{ReLU}\\
Upsampling & & & & (2 $\times$ 2) & & \\
2D Conv. & \multicolumn{1}{c}{3} & $\surd$ & (1 $\times$ 1) & (1 $\times$ 1) & $\surd$ & \multicolumn{1}{c}{Tanh}\\
\bottomrule
\end{tabular}
\end{table}

%%%%%%%%%%%%%%%%%%%%%%%%%%%%%%%%%%%%%%%%%%%%%%%%%%%%%%%%%%%%%%%%

\clearpage
\subsection*{6-2. MNIST}

\begin{table}[h]
\renewcommand{\arraystretch}{.95}
\centering
\caption{MNIST encoder $\mathcal{E}_{\theta}$ network.}
\label{tab:encoder}
\begin{tabular}{@{}rlcccccl@{}}
\toprule
Operation & Feature Maps & Batch Norm. & Kernels & Strides & Padding & Activation &\\ \midrule

Input $\mathbf{x}\in\mathbb{R}^{28 \times 28 \times 1}$ &  &  &  &  &  &  \\
2D Conv. & \multicolumn{1}{c}{32} & $\surd$ & (3 $\times$ 3) & (1 $\times$ 1) & $\surd$ & \multicolumn{1}{c}{ReLU} \\
(ENC1) 2D Conv. & \multicolumn{1}{c}{32} & $\surd$ & (4 $\times$ 4) & (2 $\times$ 2) & $\surd$ & \multicolumn{1}{c}{ReLU} \\
2D Conv. & \multicolumn{1}{c}{64} & $\surd$ & (3 $\times$ 3) & (1 $\times$ 1) & $\surd$ & \multicolumn{1}{c}{ReLU} \\
(ENC2) 2D Conv. & \multicolumn{1}{c}{64} & $\surd$ & (4 $\times$ 4) & (2 $\times$ 2) & $\surd$ & \multicolumn{1}{c}{ReLU} \\
2D Conv. & \multicolumn{1}{c}{128} & $\surd$ & (3 $\times$ 3) & (1 $\times$ 1) & $\surd$ & \multicolumn{1}{c}{ReLU} \\
(ENC3) 2D Conv. & \multicolumn{1}{c}{128} & $\surd$ & (4 $\times$ 4) & (2 $\times$ 2) & $\surd$ & \multicolumn{1}{c}{ReLU} \\
2D Conv. & \multicolumn{1}{c}{256} & $\surd$ & (3 $\times$ 3) & (1 $\times$ 1) & $\surd$ & \multicolumn{1}{c}{ReLU} \\
(ENC4) 2D Conv. & \multicolumn{1}{c}{256} & $\surd$ & (4 $\times$ 4) & (2 $\times$ 2) & $\surd$ & \multicolumn{1}{c}{ReLU} \\
Output $\mathbf{z}\in\mathbb{R}^{2 \times 2 \times 256}$ &  &  &  &  &  &  \\
\bottomrule
\end{tabular}
\end{table}

\begin{table}[h]
\renewcommand{\arraystretch}{.95}
\centering
\caption{MNIST classifier $\mathcal{F}$ network.}
\label{tab:encoder}
\begin{tabular}{@{}rlcccccl@{}}
\toprule
Operation & Feature Maps & Batch Norm. & Dropout & Activation &\\ \midrule
Input $\mathbf{z}\in\mathbb{R}^{2 \times 2 \times 256}$ &  &  &  &  &  &  \\
Fully Connected & \multicolumn{1}{c}{128} & $\times$ & 0.5 & \multicolumn{1}{c}{ReLU} \\
Fully Connected & \multicolumn{1}{c}{10} & $\times$ & 0.25 & \multicolumn{1}{c}{Softmax} \\

\bottomrule
\end{tabular}
\end{table}

\begin{table}[h]
\renewcommand{\arraystretch}{.95}
\centering
\caption{MNIST discriminator $\mathcal{D}_{\psi}$ network.}
\label{tab:encoder}
\begin{tabular}{@{}rlcccccl@{}}
\toprule
Operation & Feature Maps & Batch Norm. & Kernels & Strides & Padding & Activation &\\ \midrule

Input $\mathbf{x}\in\mathbb{R}^{28 \times 28 \times 1}$ &  &  &  &  &  &  \\
2D Conv. & \multicolumn{1}{c}{32} & $\times$ & (3 $\times$ 3) & (1 $\times$ 1) & $\surd$ & \multicolumn{1}{c}{Leaky ReLU} \\
2D Conv. & \multicolumn{1}{c}{32} & $\surd$ & (4 $\times$ 4) & (2 $\times$ 2) & $\surd$ & \multicolumn{1}{c}{Leaky ReLU} \\
2D Conv. & \multicolumn{1}{c}{64} & $\surd$ & (3 $\times$ 3) & (1 $\times$ 1) & $\surd$ & \multicolumn{1}{c}{Leaky ReLU} \\
2D Conv. & \multicolumn{1}{c}{64} & $\surd$ & (4 $\times$ 4) & (2 $\times$ 2) & $\surd$ & \multicolumn{1}{c}{Leaky ReLU} \\
2D Conv. & \multicolumn{1}{c}{128} & $\surd$ & (3 $\times$ 3) & (1 $\times$ 1) & $\surd$ & \multicolumn{1}{c}{Leaky ReLU} \\
2D Conv. & \multicolumn{1}{c}{128} & $\surd$ & (4 $\times$ 4) & (2 $\times$ 2) & $\surd$ & \multicolumn{1}{c}{Leaky ReLU} \\
2D Conv. & \multicolumn{1}{c}{256} & $\surd$ & (3 $\times$ 3) & (1 $\times$ 1) & $\surd$ & \multicolumn{1}{c}{Leaky ReLU} \\
2D Conv. & \multicolumn{1}{c}{256} & $\surd$ & (4 $\times$ 4) & (2 $\times$ 2) & $\surd$ & \multicolumn{1}{c}{Leaky ReLU} \\
Fully Connected & \multicolumn{1}{c}{1} & $\times$ & & & & \multicolumn{1}{c}{Linear}\\
\bottomrule
\end{tabular}
\end{table}

\begin{table}[h]
\renewcommand{\arraystretch}{.95}
\centering
\caption{MNIST counterfactual map generator $\mathcal{G}_{\phi}$ network.}
\label{tab:mnist_g}
\begin{tabular}{@{}rlcccccl@{}}
\toprule
Operation & Feature Maps & Batch Norm. & Kernels & Strides & Padding & Activation &\\ \midrule
Input $\mathbf{z}\in\mathbb{R}^{2 \times 2 \times 256}$ &  &  &  &  &  &  \\
Upsampling & & & & (2 $\times$ 2) & & \\
(DEC3) 2D Conv. & \multicolumn{1}{c}{128} & $\surd$ & (3 $\times$ 3) & (1 $\times$ 1) & $\surd$ & \multicolumn{1}{c}{ReLU}\\
\multicolumn{5}{c}{Concatenate $\text{(ENC3)}^*$ and (DEC3) along the channel axis} \\
2D Conv. & \multicolumn{1}{c}{128} & $\surd$ & (3 $\times$ 3) & (1 $\times$ 1) & $\surd$ & \multicolumn{1}{c}{ReLU}\\

Upsampling & & & & (2 $\times$ 2) & & \\
(DEC2) 2D Conv. & \multicolumn{1}{c}{64} & $\surd$ & (2 $\times$ 2) & (1 $\times$ 1) &  & \multicolumn{1}{c}{ReLU}\\
\multicolumn{5}{c}{Concatenate $\text{(ENC2)}^*$ and (DEC2) along the channel axis} \\
2D Conv. & \multicolumn{1}{c}{64} & $\surd$ & (3 $\times$ 3) & (1 $\times$ 1) & $\surd$ & \multicolumn{1}{c}{ReLU}\\

Upsampling & & & & (2 $\times$ 2) & & \\
(DEC1) 2D Conv. & \multicolumn{1}{c}{32} & $\surd$ & (3 $\times$ 3) & (1 $\times$ 1) & $\surd$ & \multicolumn{1}{c}{ReLU}\\
\multicolumn{5}{c}{Concatenate $\text{(ENC1)}^*$ and (DEC1) along the channel axis} \\
2D Conv. & \multicolumn{1}{c}{32} & $\surd$ & (3 $\times$ 3) & (1 $\times$ 1) & $\surd$ & \multicolumn{1}{c}{ReLU}\\
2D Deconv. & \multicolumn{1}{c}{1} & $\surd$ & (4 $\times$ 4) & (2 $\times$ 2) & $\surd$ & \multicolumn{1}{c}{Tanh}\\

\bottomrule
\end{tabular}
\end{table}

%%%%%%%%%%%%%%%%%%%%%%%%%%%%%%%%%%%%%%%%%%%%%%%%%%%%%%%%%%%%%%%%

\clearpage
\subsection*{6-3. ADNI}
\begin{table}[h]
\renewcommand{\arraystretch}{.95}
\centering
\caption{ADNI encoder $\mathcal{E}_{\theta}$ network.}
\label{tab:encoder}
\begin{tabular}{@{}rlccccccl@{}}
\toprule
Operation & Feature Maps & Batch Norm. & Kernels & Strides & Padding & Activation &\\ \midrule

Input $\mathbf{x}\in\mathbb{R}^{96 \times 114 \times 96 \times 1}$ &  &  &  &  &  &  &  \\
3D Conv. & \multicolumn{1}{c}{16} & $\surd$ & (3 $\times$ 3 $\times$ 3) & (1 $\times$ 1 $\times$ 1) & $\surd$ & \multicolumn{1}{c}{ReLU} \\
(ENC1) 3D Conv. & \multicolumn{1}{c}{16} & $\surd$ & (3 $\times$ 3 $\times$ 3) & (1 $\times$ 1 $\times$ 1) & $\surd$ & \multicolumn{1}{c}{ReLU} \\
Max Pooling & & & (2 $\times$ 2 $\times$ 2) & (2 $\times$ 2 $\times$ 2 ) &  & \\

3D Conv. & \multicolumn{1}{c}{32} & $\surd$ & (3 $\times$ 3 $\times$ 3) & (1 $\times$ 1 $\times$ 1) & $\surd$ & \multicolumn{1}{c}{ReLU} \\
(ENC2) 3D Conv. & \multicolumn{1}{c}{32} & $\surd$ & (3 $\times$ 3 $\times$ 3) & (1 $\times$ 1 $\times$ 1) & $\surd$ & \multicolumn{1}{c}{ReLU} \\
Max Pooling & & & (2 $\times$ 2 $\times$ 2) & (2 $\times$ 2 $\times$ 2) &  & \\

3D Conv. & \multicolumn{1}{c}{64} & $\surd$ & (3 $\times$ 3 $\times$ 3) & (1 $\times$ 1 $\times$ 1) & $\surd$ & \multicolumn{1}{c}{ReLU} \\
3D Conv. & \multicolumn{1}{c}{64} & $\surd$ & (3 $\times$ 3 $\times$ 3) & (1 $\times$ 1 $\times$ 1) & $\surd$ & \multicolumn{1}{c}{ReLU} \\
(ENC3) 3D Conv. & \multicolumn{1}{c}{64} & $\surd$ & (3 $\times$ 3 $\times$ 3) & (1 $\times$ 1 $\times$ 1) & $\surd$ & \multicolumn{1}{c}{ReLU} \\
Max Pooling & & & (2 $\times$ 2 $\times$ 2) & (2 $\times$ 2 $\times$ 2) &  & \\

3D Conv. & \multicolumn{1}{c}{128} & $\surd$ & (3 $\times$ 3 $\times$ 3) & (1 $\times$ 1 $\times$ 1) & $\surd$ & \multicolumn{1}{c}{ReLU} \\
3D Conv. & \multicolumn{1}{c}{128} & $\surd$ & (3 $\times$ 3 $\times$ 3) & (1 $\times$ 1 $\times$ 1) & $\surd$ & \multicolumn{1}{c}{ReLU} \\
(ENC4) 3D Conv. & \multicolumn{1}{c}{128} & $\surd$ & (3 $\times$ 3$\times$ 3) & (1 $\times$ 1 $\times$ 1) & $\surd$ & \multicolumn{1}{c}{ReLU} \\
Max Pooling & & & (2 $\times$ 2 $\times$ 2) & (2 $\times$ 2 $\times$ 2) &  & \\

3D Conv. & \multicolumn{1}{c}{128} & $\surd$ & (3 $\times$ 3 $\times$ 3) & (1 $\times$ 1 $\times$ 1) & $\surd$ & \multicolumn{1}{c}{ReLU} \\
3D Conv. & \multicolumn{1}{c}{128} & $\surd$ & (3 $\times$ 3 $\times$ 3) & (1 $\times$ 1 $\times$ 1) & $\surd$ & \multicolumn{1}{c}{ReLU} \\
3D Conv. & \multicolumn{1}{c}{128} & $\surd$ & (3 $\times$ 3 $\times$ 3) & (1 $\times$ 1 $\times$ 1) & $\surd$ & \multicolumn{1}{c}{ReLU} \\
 Output $\mathbf{z}\in\mathbb{R}^{6 \times 7 \times 6 \times 128}$ &  &  &  &  &  &  &  \\
\bottomrule
\end{tabular}
\end{table}

\begin{table}[h]
\renewcommand{\arraystretch}{.95}
\centering
\caption{ADNI classifier $\mathcal{F}$ network.}
\label{tab:encoder}
\begin{tabular}{@{}rlcccccl@{}}
\toprule
Operation & Feature Maps & Batch Norm. & Dropout & Activation &\\ \midrule
Input $\mathbf{z}\in\mathbb{R}^{6 \times 7 \times 6 \times 128}$ &  &  &  &  &  &  \\
Fully Connected & \multicolumn{1}{c}{256} & $\times$ & $\times$ & \multicolumn{1}{c}{ReLU} \\
Fully Connected & \multicolumn{1}{c}{3} & $\times$ & $\times$ & \multicolumn{1}{c}{Softmax} \\
\bottomrule
\end{tabular}
\end{table}

\begin{table}[h]
\renewcommand{\arraystretch}{.95}
\centering
\caption{ADNI discriminator $\mathcal{D}_{\psi}$ network.}
\label{tab:encoder}
\begin{tabular}{@{}rlcccccl@{}}
\toprule
Operation & Feature Maps & Batch Norm. & Kernels & Strides & Padding & Activation &\\ \midrule
Input $\mathbf{x}\in\mathbb{R}^{96 \times 114 \times 96 \times 1}$ &  &  &  &  &  &  \\
3D Conv. & \multicolumn{1}{c}{16} & $\times$ & (3 $\times$ 3 $\times$ 3) & (1 $\times$ 1 $\times$ 1) & $\surd$ & \multicolumn{1}{c}{Leaky ReLU} \\
3D Conv. & \multicolumn{1}{c}{16} & $\surd$ & (4 $\times$ 4 $\times$ 4) & (1 $\times$ 2$\times$ 2) & $\surd$ & \multicolumn{1}{c}{Leaky ReLU} \\
% Max pooling & & & (2 $\times$ 2 $\times$ 2) & (2 $\times$ 2 $\times$ 2) &  &  \\

3D Conv. & \multicolumn{1}{c}{32} & $\surd$ & (3 $\times$ 3 $\times$ 3) & (1 $\times$ 1 $\times$ 1) & $\surd$ & \multicolumn{1}{c}{Leaky ReLU} \\
3D Conv. & \multicolumn{1}{c}{32} & $\surd$ & (4 $\times$ 4 $\times$ 4) & (1 $\times$ 2 $\times$ 2) & $\surd$ & \multicolumn{1}{c}{Leaky ReLU} \\
% Max pooling & & & (2 $\times$ 2 $\times$ 2) & (2 $\times$ 2 $\times$ 2) &  & \\

3D Conv. & \multicolumn{1}{c}{64} & $\surd$ & (3 $\times$ 3 $\times$ 3) & (1 $\times$ 1 $\times$ 1) & $\surd$ & \multicolumn{1}{c}{Leaky ReLU} \\
3D Conv. & \multicolumn{1}{c}{64} & $\surd$ & (3 $\times$ 3 $\times$ 3) & (1 $\times$ 1 $\times$ 1) & $\surd$ & \multicolumn{1}{c}{Leaky ReLU} \\
3D Conv. & \multicolumn{1}{c}{64} & $\surd$ & (4 $\times$ 4 $\times$ 4) & (2 $\times$ 2 $\times$ 2) & $\surd$ & \multicolumn{1}{c}{Leaky ReLU} \\
% Max pooling & & & (2 $\times$ 2 $\times$ 2) & (2 $\times$ 2 $\times$ 2) &  & \\

3D Conv. & \multicolumn{1}{c}{128} & $\surd$ & (3 $\times$ 3 $\times$ 3) & (1 $\times$ 1 $\times$ 1) & $\surd$ & \multicolumn{1}{c}{Leaky ReLU} \\
3D Conv. & \multicolumn{1}{c}{128} & $\surd$ & (3 $\times$ 3 $\times$ 3) & (1 $\times$ 1 $\times$ 1) & $\surd$ & \multicolumn{1}{c}{Leaky ReLU} \\
3D Conv. & \multicolumn{1}{c}{128} & $\surd$ & (4 $\times$ 4 $\times$ 4) & (2 $\times$ 2 $\times$ 2) & $\surd$ & \multicolumn{1}{c}{Leaky ReLU} \\
% Max pooling & & & (2 $\times$ 2 $\times$ 2) & (2 $\times$ 2 $\times$ 2) &  & \\

3D Conv. & \multicolumn{1}{c}{256} & $\surd$ & (3 $\times$ 3 $\times$ 3) & (1 $\times$ 1 $\times$ 1) & $\surd$ & \multicolumn{1}{c}{Leaky ReLU} \\
3D Conv. & \multicolumn{1}{c}{256} & $\surd$ & (3 $\times$ 3 $\times$ 3) & (1 $\times$ 1 $\times$ 1) & $\surd$ & \multicolumn{1}{c}{Leaky ReLU} \\
3D Conv. & \multicolumn{1}{c}{256} & $\surd$ & (4 $\times$ 4 $\times$ 4) & (2 $\times$ 2 $\times$ 2) & $\surd$ &  \multicolumn{1}{c}{Leaky ReLU} \\

3D Conv. & \multicolumn{1}{c}{64} & $\surd$ & (1 $\times$ 1 $\times$ 1) & (1 $\times$ 1 $\times$ 1) & $\surd$ & \multicolumn{1}{c}{Leaky ReLU} \\
3D Conv. & \multicolumn{1}{c}{1} & $\surd$ & (1 $\times$ 1 $\times$ 1) & (1 $\times$ 1 $\times$ 1) & $\surd$ & \multicolumn{1}{c}{Linear} \\
Global Average Pooling &  &  &  &  &  & \\
\bottomrule
\end{tabular}
\end{table}

\clearpage

\begin{table}[h]
\renewcommand{\arraystretch}{.9}
\centering
\caption{ADNI counterfactual map generator $\mathcal{G}_{\phi}$ network.}
\label{tab:adni_g}
\begin{tabular}{@{}rlcccccl@{}}
\toprule
Operation & Feature Maps & Batch Norm. & Kernels & Strides & Padding & Activation &\\ \midrule
Input $\mathbf{z}\in\mathbb{R}^{6 \times 7 \times 6 \times 128}$ &  &  &  &  &  &  \\
(DEC4) Upsampling & & & & (2 $\times$ 2 $\times$ 2) & & \\
\multicolumn{5}{c}{Concatenate $\text{(ENC4)}^*$ and (DEC4) along the channel axis} \\
3D Conv. & \multicolumn{1}{c}{128} & $\surd$ & (3 $\times$ 3 $\times$ 3) & (1 $\times$ 1 $\times$ 1) & $\surd$ & \multicolumn{1}{c}{ReLU}\\
3D Conv. & \multicolumn{1}{c}{128} & $\surd$ & (3 $\times$ 3 $\times$ 3) & (1 $\times$ 1 $\times$ 1) & $\surd$ & \multicolumn{1}{c}{ReLU}\\
3D Conv. & \multicolumn{1}{c}{128} & $\surd$ & (3 $\times$ 3 $\times$ 3) & (1 $\times$ 1 $\times$ 1) & $\surd$ & \multicolumn{1}{c}{ReLU}\\

(DEC3) Upsampling & & & & (2 $\times$ 2 $\times$ 2) & & \\
\multicolumn{5}{c}{Concatenate $\text{(ENC3)}^*$ and (DEC3) along the channel axis} \\
3D Conv. & \multicolumn{1}{c}{64} & $\surd$ & (3 $\times$ 3 $\times$ 3) & (1 $\times$ 1 $\times$ 1) & $\surd$ & \multicolumn{1}{c}{ReLU}\\
3D Conv. & \multicolumn{1}{c}{64} & $\surd$ & (3 $\times$ 3 $\times$ 3) & (1 $\times$ 1 $\times$ 1) & $\surd$ & \multicolumn{1}{c}{ReLU}\\
3D Conv. & \multicolumn{1}{c}{64} & $\surd$ & (3 $\times$ 3 $\times$ 3) & (1 $\times$ 1 $\times$ 1) & $\surd$ & \multicolumn{1}{c}{ReLU}\\

(DEC2) Upsampling & & & & (2 $\times$ 2 $\times$ 2) & & \\
\multicolumn{5}{c}{Concatenate $\text{(ENC2)}^*$ and (DEC2) along the channel axis} \\
3D Deconv & \multicolumn{1}{c}{32} & $\surd$ & (1 $\times$ 2 $\times$ 1) & (1 $\times$ 1 $\times$ 1) &  & \multicolumn{1}{c}{ReLU}\\
3D Conv. & \multicolumn{1}{c}{32} & $\surd$ & (3 $\times$ 3 $\times$ 3) & (1 $\times$ 1 $\times$ 1) & $\surd$ & \multicolumn{1}{c}{ReLU}\\
3D Conv. & \multicolumn{1}{c}{32} & $\surd$ & (3 $\times$ 3 $\times$ 3) & (1 $\times$ 1 $\times$ 1) & $\surd$ & \multicolumn{1}{c}{ReLU}\\

(DEC1) Upsampling & & & & (2 $\times$ 2 $\times$ 2) & & \\
\multicolumn{5}{c}{Concatenate $\text{(ENC1)}^*$ and (DEC1) along the channel axis} \\
3D Conv. & \multicolumn{1}{c}{16} & $\surd$ & (3 $\times$ 3 $\times$ 3) & (1 $\times$ 1 $\times$ 1) & $\surd$ & \multicolumn{1}{c}{ReLU}\\
3D Conv. & \multicolumn{1}{c}{16} & $\surd$ & (3 $\times$ 3 $\times$ 3) & (1 $\times$ 1 $\times$ 1) & $\surd$ & \multicolumn{1}{c}{ReLU}\\
3D Conv. & \multicolumn{1}{c}{1} & $\surd$ & (1 $\times$ 1 $\times$ 1) & (1 $\times$ 1 $\times$ 1) & $\surd$ & \multicolumn{1}{c}{Linear}\\
\bottomrule
\end{tabular}
\end{table}

\subsection*{6-4. Hyperparameters}

Best performing model hyperparameters are shown below. $\lambda_2 =0$ for 3D Shapes and MNIST since $\ell_2$ normalization tend to soften the edges, which is beneficial for ADNI counterfactual maps but disadvantageous for images with hard edges (\ie, 3D Shapes, MNIST).

\begin{table}[h]
\renewcommand{\arraystretch}{.8}
\centering
\caption{3D Shapes model hyperparameters.}
\label{tab:omni-hyper}
\begin{tabular}{@{}rlccl@{}}
\toprule
Optimizer & \multicolumn{4}{l}{Adam $\left(\beta_1=0.9, \beta_2=0.999\right)$} \\
Epochs & \multicolumn{4}{l}{50}\\
Batch Size & \multicolumn{4}{l}{128} \\
Learning Rate & \multicolumn{4}{l}{$\text{Gen}=0.001, \text{Dis}=0.001$}\\
Exponential Decay Rate & \multicolumn{4}{l}{0.98}\\
One-sided Label Smoothing & \multicolumn{4}{l}{0.1}\\
Weight Constants & \multicolumn{4}{l}{$\lambda_1=1, \lambda_2=0, \lambda_3=5, \lambda_4=10, \lambda_5=1, \lambda_6=1$} \\
\bottomrule
\end{tabular}
\end{table}

\begin{table}[h]
\renewcommand{\arraystretch}{.8}
\centering
\caption{MNIST model hyperparameters.}
\label{tab:omni-hyper}
\begin{tabular}{@{}rlccl@{}}
\toprule
Optimizer & \multicolumn{4}{l}{Adam $\left(\beta_1=0.9, \beta_2=0.999\right)$} \\
Epochs & \multicolumn{4}{l}{100}\\
Batch Size & \multicolumn{4}{l}{256} \\
Learning Rate & \multicolumn{4}{l}{$\text{Gen}=0.001, \text{Dis}=0.001$}\\
Exponential Decay Rate & \multicolumn{4}{l}{0.99}\\
One-sided Label Smoothing & \multicolumn{4}{l}{$\times $}\\
Weight Constants & \multicolumn{4}{l}{$\lambda_1=1, \lambda_2=0, \lambda_3=0.5, \lambda_4=1, \lambda_5=1, \lambda_6=1$} \\
\bottomrule
\end{tabular}
\end{table}

\begin{table}[h]
\renewcommand{\arraystretch}{.8}
\centering
\caption{ADNI model hyperparameters.}
\label{tab:omni-hyper}
\begin{tabular}{@{}rlccl@{}}
\toprule
Optimizer & \multicolumn{4}{l}{Adam $\left(\beta_1=0.9, \beta_2=0.999\right)$} \\
Epochs & \multicolumn{4}{l}{100}\\
Batch Size & \multicolumn{4}{l}{3} \\
Learning Rate & \multicolumn{4}{l}{$\text{Gen}=0.01, \text{Dis}=0.01$}\\
Exponential Decay Rate & \multicolumn{4}{l}{1.0}\\
One-sided Label Smoothing & \multicolumn{4}{l}{$\times$}\\
Weight Constants & \multicolumn{4}{l}{$\lambda_1=1, \lambda_2=10, \lambda_3=5, \lambda_4=10, \lambda_5=1, \lambda_6=1$} \\
\bottomrule
\end{tabular}
\end{table}

%%%%%%%%%%%%%%%%%%%%%%%%%%%%%%%%%%%%%%%%%%%%%%%%%%%%%%%%%%%%%%%%
\clearpage